\documentclass[review,onefignum,onetabnum, noline]{siamonline250211}

\usepackage{graphicx} 
\usepackage[utf8]{inputenc}
\usepackage{graphicx}
\usepackage{diagbox} 
\usepackage{hyperref}
\usepackage{amsmath}
\usepackage{paralist}
\usepackage{amsfonts}
\usepackage{amssymb} 
\usepackage{algorithm}
\usepackage{algorithmic}
\usepackage{harpoon}
\usepackage{float}
\usepackage{booktabs}
\usepackage{changes}
\usepackage{url}
\usepackage{multirow}
\usepackage{comment}
\usepackage{mathtools,nccmath}
\usepackage{bm}
\renewtheorem{theorem}{Theorem}[section]
\newtheorem{example}{Example}[section]
\renewtheorem{lemma}{Lemma}[section]
\renewtheorem{corollary}{Corollary}[section]
\newtheorem{assumption}{Assumption}[section]
\renewtheorem{definition}{Definition}[section]

\usepackage{hyperref}

\def\d{\text{d}}
\def\E{\mathbb{E}}

\usepackage{lipsum}
\usepackage{amsfonts}
\usepackage{graphicx}
\usepackage{epstopdf}
\usepackage{hyperref}
\usepackage{algorithmic}
\ifpdf
  \DeclareGraphicsExtensions{.eps,.pdf,.png,.jpg}
\else
  \DeclareGraphicsExtensions{.eps}
\fi

\usepackage{enumitem}
\setlist[enumerate]{leftmargin=.5in}
\setlist[itemize]{leftmargin=.5in}

\newsiamremark{remark}{Remark}
\newsiamremark{hypothesis}{Hypothesis}
\crefname{hypothesis}{Hypothesis}{Hypotheses}
\newsiamthm{claim}{Claim}
\newsiamremark{fact}{Fact}
\crefname{fact}{Fact}{Facts}
\usepackage{hyperref}

\headers{A generalized Wasserstein-2 distance approach}{M. Xia, Q. Shen}

\title{A generalized Wasserstein-2 distance approach for efficient reconstruction of random field models using stochastic neural networks\thanks{Submitted to the editors 07/07/2025.
\funding{This work was supported in part through the NYU IT High Performance Computing resources, services, and staff expertise.}}}

\author{
  Mingtao Xia$^{*, }$\footnotemark[2]
  \and
  Qijing Shen\footnotemark[3]
}

\usepackage{url}
\footnotetext[1]{Department of Mathematics, University of Houston, Houston, TX (\texttt{mxia4@uh.edu}).}
\footnotetext[2]{Courant Institute of Mathematical Sciences, New York University, New York, NY.}
\footnotetext[3]{Nuffield School of Medicine, University of Oxford, Oxford, UK (\texttt{qijing.shen@reuben.ox.ac.uk}).}

\usepackage{amsopn}

\ifpdf
\hypersetup{
  pdftitle={A generalized Wasserstein-2 distance approach for efficient reconstruction of random field models using stochastic neural networks},
  pdfauthor={Mingtao Xia, Qijing Shen}
}
\fi
\nolinenumbers

\begin{document}

\maketitle

\begin{abstract}
    In this work, we propose a novel generalized Wasserstein-2 distance approach for efficiently training stochastic neural networks to reconstruct random field models, where the target random variable comprises both continuous and categorical components. We prove that a stochastic neural network can approximate random field models under a Wasserstein-2 distance metric
    under nonrestrictive conditions. Furthermore, this stochastic neural network can be efficiently trained by minimizing our proposed generalized local squared Wasserstein-2 loss function. We showcase the effectiveness of our proposed approach in various uncertainty quantification tasks, including classification, reconstructing the distribution of mixed random variables, and learning complex noisy dynamical systems from spatiotemporal data. 
\end{abstract}



\section{Introduction}
Random field models, in which the outcome is a random variable whose distribution is determined by observed features, 
have found wide applications across different fields. 
For example, in engineering, reliability analysis and signal processing require taking into account the randomness of the output given the input \cite{nowak2012reliability}. In medical fields, randomized controlled trials also rely on probabilistic designs \cite{friedman2015fundamentals}. Additionally, it is necessary to take into account the stochasticity in customers' choices and preferences for economics \cite{mcfadden1974conditional}.

Reconstruction of the distribution of the target random variable from a finite number of observed data is receiving increasing research interest in uncertainty quantification (UQ) and related fields.
For reconstructing the distribution of categorical random variables, common approaches include multinomial logistic regression \cite{Agresti2018} and Bayesian network modeling \cite{Koller2009}. Continuous variables are often analyzed using linear models \cite{Hastie2009} or nonparametric density estimation \cite{Silverman1986}. Many real-world applications also involve the reconstruction of distributions of mixed variables containing both continuous and discrete components. To handle such cases, generalized linear mixed models \cite{McCulloch2008} and latent variable approaches \cite{Bartholomew2011} provide flexible frameworks. Additionally, Bayesian nonparametric methods \cite{Ghahramani2013} offer additional ways to model the dependence of mixed random variables on given features.

The Wasserstein distance, also known as the earth mover’s distance, has emerged as a powerful tool for comparing probability distributions \cite{villani2009optimal, zheng2020nonparametric}, particularly in UQ fields involving noisy data. 
For example, in computational biology, Wasserstein metrics help compare cell population distributions, particularly in single-cell transcriptomics \cite{schiebinger2019optimal}. Furthermore, in image processing and shape analysis, the Wasserstein distance is effective in comparing histograms and distributions with spatial structure \cite{solomon2015convolutional}. Additionally, in machine learning, the Wasserstein generative adversarial network (WGAN) has found wide applications in different tasks, such as image generation \cite{jin2019image, wang2023improved} and generating the distribution of solutions to partial differential equations with latent parameters \cite{gao2022wasserstein}. 

Recently, direct minimization of the Wasserstein distance as a loss function
to train neural networks has been investigated for multiple UQ tasks. For example, in \cite{xia2024efficient, xia2024squared}, a temporally decoupled squared Wasserstein-2 ($W_2$) distance loss function has been proposed for reconstructing different stochastic processes.
In \cite{xia2024local}, a local squared $W_2$ method has been proposed to efficiently train a stochastic neural network (SNN) for the reconstruction of random functions. However, for categorical random variables, the Wasserstein distance is not directly applicable as there is usually not a ``distance" for categorical variables. Additionally, the ``discrete randomness" issue may pose substantial difficulty for automatic differentiation when the target variable is categorical \cite{arya2022automatic}.

In this work, given a probability space $(\Omega, \mathcal{F}, P)$, we develop a novel generalized Wasserstein distance method to reconstruct a random field model: 
\begin{equation}
    \bm{y}_{\bm{x}}\coloneqq \bm{y}(\bm{x};\omega),\,\, \bm{x}\in D\subseteq\mathbb{R}^n,
    \label{random_field_model}
\end{equation}
where $\bm{x}\in D$ denotes continuous features, $D$ is a bounded set in $\mathbb{R}^n$, and $\omega\in\Omega$ is the latent random variable. We use a stochastic neural network (SNN) to approximate Eq.~\eqref{random_field_model}, whose output is another random field model:
\begin{equation}
    \hat{\bm{y}}_{\bm{x}}\coloneqq \bm{y}(\bm{x};\hat{\omega}),\,\, \bm{x}\in D\subseteq\mathbb{R}^n,
    \label{approximate_random_field}
\end{equation}
where $\hat{\omega}\in\hat{\Omega}$ denotes the random variable in the SNN model (the sample space $\hat{\Omega}$ does not need to be the same as $\Omega$). 
For each $\bm{x}$, 
\begin{equation}
\bm{y}_{\bm{x}}\coloneqq(y_1(\bm{x};\omega),...,y_{d_1}(\bm{x};\omega), y_{d_1+1}(\bm{x};\omega), ..., y_{n}(\bm{x};\omega))
\label{random_variable}
\end{equation}
and 
\begin{equation}
\hat{\bm{y}}_{\bm{x}}\coloneqq(\hat{y}_1(\bm{x};\hat{\omega}),...,\hat{y}_{d_1}(\bm{x};\hat{\omega}), \hat{y}_{d_1+1}(\bm{x};\hat{\omega}), ..., \hat{y}_{n}(\bm{x};\hat{\omega}))
\label{approximate_random_variable}
\end{equation}
are both $d$-dimensional random variable. Specifically, in Eq.~\eqref{random_field_model}
$y_1,...,y_{d_1}$ are continuous and $y_{d_1+1},...,y_{d}$ are categorical.

\subsection{Our contributions}
In this work, we proposed a generalized $W_2$ method for the reconstruction of the random field model Eq.~\eqref{random_field_model}. Our contributions are as follows:
\begin{enumerate}
        \item We propose a 
        generalized $W_2$ distance approach for training SNNs to reconstruct the random field model Eq.~\eqref{random_field_model} where $\bm{y}_{\bm{x}}$ is a mixed random variable. Specifically, we proved a universal approximation property of the SNN model for approximating Eq.~\eqref{random_field_model} under this generalized $W_2$ distance metric.
        \item We develop a differentiable generalized local squared $W_2$ loss function, which can be minimized to directly train SNNs to reconstruct the random field model Eq.~\eqref{random_field_model}.
    \item We successfully apply our approach to different UQ tasks, including classification, reconstructing the distribution of mixed random variables, and learning complex noisy dynamical systems. 
\end{enumerate}

\subsection{Paper organization}
The organization of this paper is as follows: in Section~\ref{generalized_w2}, we introduce and analyze the generalized $W_2$ method for training SNNs to reconstruct the random field Eq.~\eqref{random_field_model}. In Section~\ref{numerical_results}, we test our proposed method on various UQ tasks and benchmark it against other UQ methods. In Section~\ref{conclusion}, we summarize our results and discuss potential future directions. Notations and symbols that are often used throughout this paper are summarized in Table~\ref{tab:notation}.

\begin{table}[h]
\centering
\footnotesize
\begin{tabular}{ll}
\toprule
\textbf{Symbol} & \textbf{Description} \\
\midrule
\( \bm{x} \)         & Input variable (features) in $\mathbb{R}^n$. \\
\( \bm{y}_{\bm{x}} \)         & Target random variable in the ground-truth uncertainty model Eq.~\eqref{random_variable} in $\mathbb{R}^d$. \\
\( \hat{\bm{y}}_{\bm{x}} \)         & Output of the approximate uncertainty model Eq.~\eqref{approximate_random_variable} in $\mathbb{R}^d$. \\
$\delta$ & The size of the neighborhood for $\bm{x}$.  \\
$N$ & The number of total training samples.\\
$N(\bm{x},\delta)$ & The number of samples $(\bm{x}_i, \bm{y}_i)$ satisfying $\|\bm{x}_i - \bm{x}\|_{2}\leq \delta$. $\|\cdot\|_2$ is the $\ell^2$ norm for $\bm{x}\in\mathbb{R}^n$.\\
\( f_{\bm{x}} \) ($\hat{f}_{\bm{x}}$)       & The probability measure of $\bm{y}(\bm{x};\omega)$ ($\hat{\bm{y}}(\bm{x};\hat{\omega})$) given $\bm{x}$. \\
\( f_{\bm{x}, \delta}^{\text{e}} \) ($\hat{f}_{\bm{x}, \delta}^{\text{e}}$)       & The empirical probability measure of $\bm{y}_{\tilde{\bm{x}}}$ ($\hat{\bm{y}}_{\tilde{\bm{x}}}$) conditioned on $\|\tilde{\bm{x}}-\bm{x}\|_2\leq\delta$. \\
\(\pi_{f,  \hat{f}}\) & A coupling measure of $f$ and $\hat{f}$ whose marginal distributions coincide with $f$ and $\hat{f}$.\\
\( \hat{W}_2(f, \hat{f}) \) & The generalized Wasserstein-2 distance between two probability measures \( f \) and \( \hat{f} \). \\
$\hat{W}^{2}_{2}(\bm{y}_{\bm{x}}, \hat{\bm{y}}_{\bm{x}})$ & The squared generalized $W_2$ distance between two random fields $\bm{y}_{\bm{x}}$ and $\hat{\bm{y}}_{{\bm{x}}}$\\
& (Defined in Definition~\ref{local_w2}).\\
$\hat{W}^{2, \text{e}}_{2, \delta}(\bm{y}_{\bm{x}}, \hat{\bm{y}}_{\bm{x}})$ & The generalized local squared $W_2$ loss function.\\
\bottomrule
\end{tabular}
\vspace{0.05in}
\caption{{Summary of commonly used notations and symbols throughout the paper.}}
\label{tab:notation}
\end{table}

\section{A generalized $W_2$ method for training SNNs to reconstruct random fields}
\label{generalized_w2}
In this section, we propose our generalized $W_2$ method to train SNNs for reconstructing the random field Eq.~\eqref{random_field_model} from a finite number of observed data. 
First, we define the following norm for $\bm{y}=(y_1,...,y_d), \in\mathbb{R}^d$:
\begin{equation}
    \|\bm{y}\|^2\coloneqq \lambda \sum_{i=1}^{d_1} y_i^2 + \sum_{i=d_1+1}^{n} \hat{\delta}_{y_j, 0},
    \label{w2_distance}
\end{equation}
where $\hat{\delta}_{y_j, 0}$ is defined as
\begin{equation}
\begin{aligned}
        \hat{\delta}_{y_j, 0}=\begin{cases}
        4y_j^2, |y_j|\leq \frac{1}{2},\\
        1,\,\, |y_j|>\frac{1}{2}.
    \end{cases}
\end{aligned}
\label{delta_function}
\end{equation}
Specifically, when the last $d-d_1$ components of $\bm{y}_{\bm{x}}$  in Eqs.~\eqref{random_field_model} are all categorical, $\hat{\delta}$ becomes the Kronecker delta function. The hyperparameter $\lambda$ in Eq.~\eqref{w2_distance} signifies the weight of the continuous components $(y_1,...,y_{d_1})$ compared to the discrete components $(y_{d_1+1},...,y_d)$. As an intuitive choice, we can set $\lambda=\sum_{i=1}^{d_1}\text{Var}[y_i]$, where $\text{Var}[y_i]$ refers to data variance in the component $y_i$. The coefficient $4$ in the first line of~\eqref{delta_function} may be replaced with other constants, yet the resulting $\hat{\delta}_{y_j, 0}$ is not continuous and $\|\cdot\|$ in Eq.~\eqref{w2_distance} might not be a norm. We test how replacing the coefficient 4 with other constants in the Eq.~\eqref{delta_function} could influence the reconstruction accuracy of a random field model in Example~\ref{example1}. 

Using the distance defined in Eq.~\eqref{w2_distance}, we can defined the generalized $W_2$ distance between the probability distributions associated with $\bm{y}_{\bm{x}}$ and $\hat{\bm{y}}_{\bm{x}}$ in Eq.~\eqref{random_variable} and ~\eqref{approximate_random_variable}.
\begin{definition}
\rm 
\label{def:W2}
For $\bm{y}_{\bm{x}}, \hat{\bm{y}}_{\bm{x}}\in\mathbb{R}^n$ defined in Eq.~\eqref{random_variable} and ~\eqref{approximate_random_variable}, we assume that
\begin{equation}
    \E[\|\bm{y}_{\bm{x}}\|^2]< \infty,\,\,\,\,\E[\|\hat{\bm{y}}_{\bm{x}}\|^2]< \infty,\,\,\forall \bm{x}\in D
\end{equation}
where $\|\cdot\|$ is a distance metric defined for $\bm{y}$. We denote the probability measures associated with $\bm{y}_{\bm{x}}$ and $\hat{\bm{y}}_{\bm{x}}$ by $f_{\bm{x}}$ and $\hat{f}_{\bm{x}}$, respectively. We define the \textbf{generalized} $\bm{W_2}$ \textbf{distance}:
\begin{equation}
\hat{W}_{2}(f_{\bm{x}}, \hat{f}_{\bm{x}}) \coloneqq \inf_{{\pi_{f_{\bm{x}}, \hat f_{\bm{x}}}}}
\E_{(\bm{y}_{\bm{x}}, \hat{\bm
{y}}_{\bm{x}})\sim {\pi_{f_{\bm{x}}, \hat f_{\bm{x}}}}(\bm{y}_{\bm{x}}, \hat{\bm
{y}}_{\bm{x}})}\big[\|{\bm{y}}_{\bm{x}} - \hat{{\bm{y}}}_{\bm{x}}\|^{2}\big]^{\frac{1}{2}}.
\label{pidef}
\end{equation}
In Eq.~\eqref{pidef}, {$\pi_{f_{\bm{x}}, \hat f_{\bm{x}}}(\bm{y}_{\bm{x}}, \hat{\bm
{y}}_{\bm{x}})$ is a special coupled measure of the joint random variable $(\bm{y}_{\bm{x}}, \hat{\bm
{y}}_{\bm{x}})$, defined by the condition:}
\begin{equation}
\begin{aligned}
\begin{cases}
{{\pi_{f_{\bm{x}}, \hat f_{\bm{x}}}}}\left((A_1, A_2) \times  (\mathbb{R}^{d_1}\times S_{d-d_1})\right) =\sum_{\bm{y}_2\in A_2}\int_{A_1}{f}_{\bm{x}}(\bm{y}_1, \bm{y}_2)\d \bm{y}_1,\\
{{\pi_{f_{\bm{x}}, \hat f_{\bm{x}}}}}\left(( \mathbb{R}^{d_1}\times S_{d-d_1} )\times (A_1, A_2)\right) = \sum_{\bm{y}_2\in A_2}\int_{A_1}\hat{f}_{\bm{x}}(\bm{y}_1, \bm{y}_2)\d \bm{y}_1, 
\end{cases}\\\hspace{1.6cm}\forall (A_1, A_2)\in \mathcal{B}( \mathbb{R}^{d_1} \times S_{d-d_1}),
\end{aligned}
\label{pi_def}
\end{equation}
{where $\mathcal{B}(\mathbb{R}^{d_1}\times S_{d-d_1})$ denotes the Borel $\sigma$-algebra associated with $\mathbb{R}^{d_1}\times S_{d-d_1}$, $S_{d-d_1}\subseteq \mathbb{N}^{d-d_1}$ is a bounded set which defines all possible outcomes of the categorical components $\bm{y}_2\coloneqq(y_{d_1+1},...,y_d)$, and the infimum in Eq.~\eqref{pidef} iterates over all coupled distributions $\pi_{f_{\bm{x}}, \hat f_{\bm{x}}}(\bm{y}_{\bm{x}}, \hat{\bm
{y}}_{\bm{x}})$ of $(\bm{y}_{\bm{x}}, \hat{\bm{y}}_{\bm{x}})$ satisfying Eq.~\eqref{pi_def}.}
\end{definition}

Throughout this paper, we make the following assumptions to facilitate our analysis. 
\begin{assumption}
\rm
\label{assumptions_w2}

\begin{enumerate}
    \item We assume that $\bm{y}_{\bm{x}}$ and $\hat{\bm{y}}_{\bm{x}}$ in Eqs.~\eqref{random_variable} and \eqref{approximate_random_variable} are uniformly bounded such that there exists $0<M<\infty$:
\begin{equation}
    \max(\|\bm{y}\|, \|\bm{y}\|_2)\leq \sqrt{M}, \,\,\,\, \max(\|\bm{y}\|, \|\hat{\bm{y}}\|_2)\leq \sqrt{M}.
    \label{upperboundy}
\end{equation}
In this work, $\|\cdot\|_2$ denotes the $\ell^2$ norm of a vector in $\mathbb{R}^d$ and we have 
\begin{equation}
    \|\bm{y}\|\leq \max(2, \sqrt{\lambda})\|\bm{y}\|_2.
    \label{bounded1}
\end{equation}
\item In Eqs.~\eqref{random_variable} and \eqref{approximate_random_variable}, $\omega$ is independent of $\bm{x}$ and $\hat{\omega}$ is independent of $\bm{x}$. 
\item The probability measures associated with the mixed random variable $\bm{y}_{\bm{x}}$ in Eq.~\eqref{random_field_model} is uniform Lipschitz on $\bm{x}$ continuous
in the generalized $W_2$ distance sense: 
\begin{equation}
    \hat{W}_2(f_{\bm{x}}, f_{\tilde{\bm{x}}})\leq L \|\bm{x}-\tilde{\bm{x}}\|_2, \,\,  \forall \bm{x}, \hat{\bm{x}}\in D, 
    \label{l_condition}
\end{equation}
where $f_{\bm{x}}$ is the probability measure associated with $\bm{y}_{\bm{x}}$.
\item The probability measures associated with the mixed random variable $\hat{\bm{y}}_{\bm{x}}$ in Eq.~\eqref{approximate_random_field} is also uniform Lipschitz on $\bm{x}$ continuous
in the generalized $W_2$ distance sense: 
\begin{equation}
    \hat{W}_2(\hat{f}_{\bm{x}}, \hat{f}_{\tilde{\bm{x}}})\leq L \|\bm{x}-\tilde{\bm{x}}\|_2, \,\,  \forall \bm{x}, \hat{\bm{x}}\in D, 
    \label{l_condition2}
\end{equation}
where $\hat{f}_{\bm{x}}$ is the probability measure associated with $\hat{\bm{y}}_{\bm{x}}$.
\item For every $\bm{x}\in D$, \begin{equation}
         |f_{\bm{x}}|_{\text{mix}}\coloneqq \sum_{|\bm{n}|_0\leq d_1}\|\partial_{\bm{n}}^{|\bm{n}|_0}f_{\bm{x}}\|_{L^2}<\infty,\,\,\,|\sqrt{f_{\bm{x}}}|_{\text{mix}}<\infty
     \end{equation}
     where $|\bm{n}|_0$ is the number of nonzero components in $\bm{n}$, $\bm{n}=(n_1,...,n_j)$ satisfying $1\leq n_1<...<n_j\leq d_1$, $\|\cdot\|_{L^2}$ is the $L^2$ norm of a function, and  $\partial_{\bm{n}}f_{\bm{x}}\coloneqq\partial_{y_{n_1}}...\partial_{y_{n_j}}f$. 
     \item $|f_{\bm{x}}y_i^2|_{\text{mix}}<\infty$ and $|f_{\bm{x}}y_i^2y_j^2|_{\text{mix}}<\infty$ for $i, j=1,...,d_1$. 
     \item For every $\bm{x}\in D$, the probability measure $f_{\bm{x}}(\bm{y}_{\bm{x}})$ is uniformly continuous in the first $d_1$ continuous components of $\bm{y}_{\bm{x}}$.
\end{enumerate}
\end{assumption}

We use the notation $W_2(f_{\bm{x}}, \hat{f}_{\bm{x}})$ to denote the commonly used $W_2$ distance:
\begin{equation}
    W_2(f_{\bm{x}}, \hat{f}_{\bm{x}})\coloneqq\inf_{{\pi_{f_{\bm{x}}, \hat f_{\bm{x}}}}}
\E_{(\bm{y}_{\bm{x}}, \hat{\bm
{y}}_{\bm{x}})\sim {\pi_{f_{\bm{x}}, \hat f_{\bm{x}}}}(\bm{y}_{\bm{x}}, \hat{\bm
{y}}_{\bm{x}})}\big[\|{\bm{y}}_{\bm{x}} - \hat{{\bm{y}}}_{\bm{x}}\|_2^{2}\big]^{\frac{1}{2}},
\end{equation}
 where $\pi_{f_{\bm{x}}, \hat f_{\bm{x}}}$ is the coupling probability measure whose marginal distributions coincide with $f_{\bm{x}}$ and $\hat{f}_{\bm{x}}$, respectively. 
 Using Eq.~\eqref{bounded1}, it is easy to verify that there exists a constant $0<K<\infty$ such that:
\begin{equation}
     \hat{W}_2(f_{\bm{x}}, \hat{f}_{\bm{x}})\leq K{W}_2(f_{\bm{x}}, \hat{f}_{\bm{x}}).
    \label{upper_lower_bounds}
\end{equation} 
Furthermore, from Eq.~\eqref{upperboundy} in Assumption~\ref{assumptions_w2}, there exists another constant $0<k<\infty$ such that:
\begin{equation}
    k{W}_2(f_{\bm{x}}, \hat{f}_{\bm{x}})\leq \hat{W}_2(f_{\bm{x}}, \hat{f}_{\bm{x}}).
    \label{upper_lower_bounds2}
\end{equation}

For any coupling measure $\pi_{f_{\bm{x}}, \hat{f}_{\bm{x}}}$ of $\bm{y}_{\bm{x}}$ and $\hat{\bm{y}}_{\bm{x}}$ whose marginal distributions coincide with $f_{\bm{x}}$ and $\hat{f}_{\bm{x}}$, we have:
\begin{equation}
    \E_{(\bm{y}_{\bm{x}}, \hat{\bm{y}}_{\bm{x}})\sim\pi_{f_{\bm{x}}, \hat{f}_{\bm{x}}}}\big[\|\bm{y}_{\bm{x}}-\hat{\bm{y}}_{\bm{x}}\|^2\big]= \E_{(\bm{y}_{\bm{x}}, \hat{\bm{y}}_{\bm{x}})\sim\pi_{f_{\bm{x}}, \hat{f}_{\bm{x}}}}\Big[\sum_{i=1}^{d_1}(y_i-\hat{y}_i)^2\Big] + \sum_{i=d_1+1}^d \E_{(\bm{y}_{\bm{x}}, \hat{\bm{y}}_{\bm{x}})\sim\pi_{f_{\bm{x}}, \hat{f}_{\bm{x}}}}\big[\hat{\delta}_{{y}_i, \hat{y}_i}\big],
\end{equation}
where $y_i, \hat{y}_i$ are the $i^{\text{th}}$ components of $\bm{y}_{\bm{x}}$ and $\hat{\bm{y}}_{\bm{x}}$, respectively.
When both $y_i, \hat{y}_i\in\mathbb{Z}$ for $i=d_1+1,...,d$, we have:
\begin{equation}
    \E_{(\bm{y}_{\bm{x}}, \hat{\bm{y}}_{\bm{x}})\sim\pi_{f_{\bm{x}}, \hat{f}_{\bm{x}}}}\big[\hat{\delta}_{{y}_i, \hat{{y}}_i}\big]\geq 1 - \sum_{k\in\mathbb{Z}}\mathbb{I}_{\bm{y}_i = \hat{\bm{y}}_i=k} \geq 1 - \sum_{k\in\mathbb{Z}}\min(p_{i, k}, \hat{p}_{i, k}),
\end{equation}
where $p_{i, k} \coloneqq P(y_i=k)$, $\hat{p}_{i, k}\coloneqq P(\hat{y}_i=k)$, and $\mathbb{I}$ is the indicator function. Denoting the marginal probability densities of $(y_1(\bm{x};\omega),...,y_{d_1}(\bm{x};\omega))$ and $(\hat{y}_1(\bm{x};\hat{\omega}),...,\hat{y}_{d_1}(\bm{x};\hat{\omega}))$ by $f_{1, \bm{x}}$ and $\hat{f}_{1, \bm{x}}$, we have the following lower bound:
\begin{equation}
    \E_{(\bm{y}_{\bm{x}}, \hat{\bm{y}}_{\bm{x}})\sim\pi_{f_{\bm{x}}, \hat{f}_{\bm{x}}}}\big[\|\bm{y}- \hat{\bm{y}}\|^2\big]\geq W_2^2(f_{1, \bm{x}}, \hat{f}_{1, \bm{x}}) + \sum_{i=d_1+1}^{d}\big(1 - \sum_{k\in\mathbb{Z}}\min(p_{i, k}, \hat{p}_{i, k})\big).
\end{equation}

Taking the infimum over all coupling probability measures $\pi_{f_{\bm{x}}, \hat{f}_{\bm{x}}}$, we conclude that:
\begin{equation}
    \hat{W}_2^2(f_{\bm{x}}, \hat{f}_{\bm{x}})\geq W_2^2(f_{1, \bm{x}}, \hat{f}_{1, \bm{x}}) + \sum_{i=d_1+1}^{d}\big(1 - \sum_{k\in\mathbb{Z}}\min(p_{i, k}, \hat{p}_{i, k})\big),
    \label{lower_bound}
\end{equation}
Therefore, when the generalized $W_2$ distance $\hat{W}_2(f_{\bm{x}}, \hat{f}_{\bm{x}})$ is sufficiently small, $W_2(f_{1, \bm{x}}, \hat{f}_{1, \bm{x}})$ is small, which indicates that the marginal distribution $f_{1, \bm{x}}$ should be matched well by the marginal distribution $\hat{f}_{1, \bm{x}}$; furthermore, the marginal distribution of $\hat{y}_{j}$ should align well with the distributions of $y_{j}$ for $j=d_1+1,...,d$.

\subsection{Universal approximation ability of SNNs to approximate the random field model Eq.~\eqref{random_field_model}}
We consider using the following SNN whose output is referred to as $\hat{\bm{y}}(\bm{x};\hat{\omega})$ in Eq.~\eqref{approximate_random_field} given the input $\bm{x}$ to approximate the random field model Eq.~\eqref{random_field_model}. 
    \begin{figure}
    \centering
\includegraphics[width=0.9\linewidth]{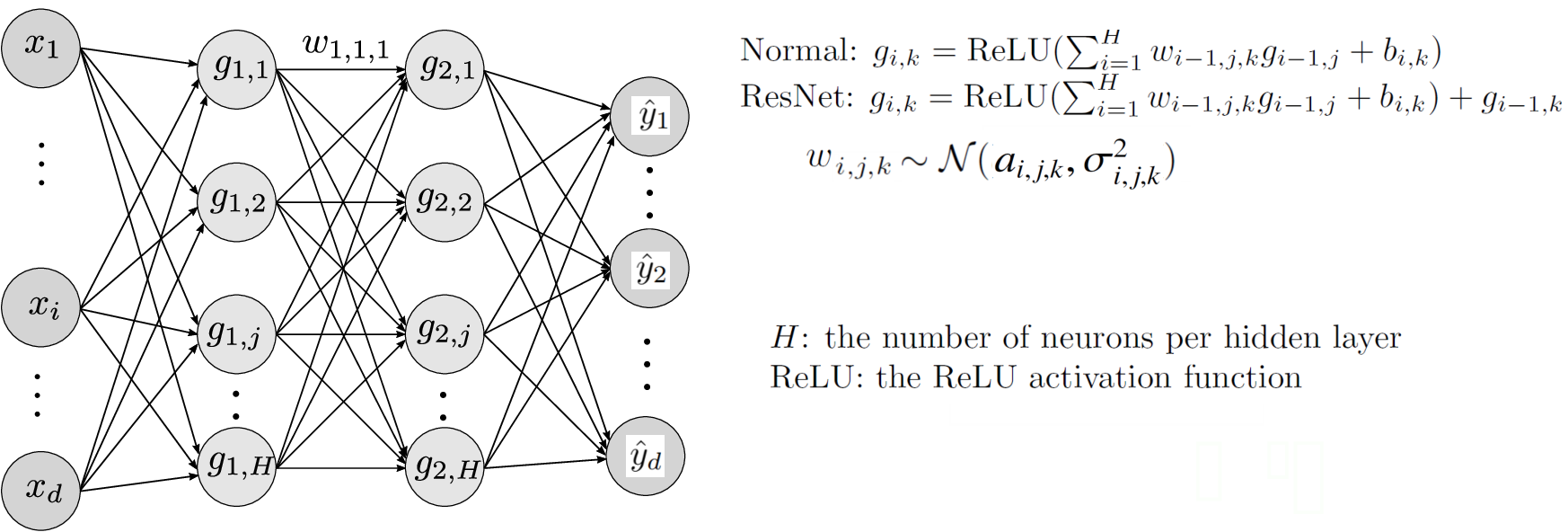}
    \caption{An example of the structure of the neural network model used in this study. In the neural network model, for each input $x$, the weights $w_{i, j, k}\sim\mathcal{N}(a_{i, j, k}, \sigma_{i, j, k}^2)$ are independently sampled. ReLU means the \texttt{ReLU} activation function and may be replaced with other activation functions. Two structures of forward propagation may be used: the normal feedforward structure or the Resnet \cite{he2016deep} structure. Note that the outputs of the SNN model are all continuous, and a rounding operation in Eq.~\eqref{rounding} is used to transform the original output $\bm{y}_{\bm{x}}$ into $\tilde{\bm{y}}_{\bm{x}}$ whose last $d-d_1$ components are categorical.}
    \label{fig:snn}
\end{figure}

We can show that, under some nonrestrictive conditions, the SNN model has the capability of approximating the random field model Eq.~\eqref{random_field_model} up to any accuracy under the $\hat{W}_2$ metric. We prove the following theorem.
\begin{theorem}
\rm
\label{theorem1}
For any random field model defined in Eq.~\eqref{random_field_model} and any positive number $\epsilon_1>0$, there exists an SNN whose output is $\hat{\bm{y}}_{\bm{x}}$ and the squared \textbf{generalized} $\bm{W_2}$ \textbf{distance between the two random fields $\bm{y}_{\bm{x}}$ and $\hat{\bm{y}}_{\bm{x}}$} satisfies:
\begin{equation}
    \hat{W}_2^2(\bm{y}_{\bm{x}},\hat{\bm{y}}_{\bm{x}})\coloneqq \int_D\hat{W}_2^2(f_{\bm{x}}, \hat{f}_{\bm{x}})\nu(\d\bm{x})\leq\epsilon_1,
    \label{generalized_define}
\end{equation}
where $f_{\bm{x}}$ is the probability measure associated with $\bm{y}(\bm{x};\omega)$ and $\hat{f}_{\bm{x}}$ is the probability measure associated with $\hat{\bm{y}}(\bm{x};\hat{\omega})$.
\end{theorem}

We prove Theorem~\ref{theorem1} in Appendix~\ref{appendixA}. From Theorem~\ref{theorem1}, any random field model in Eq.~\eqref{random_field_model} can be approximated by an SNN described in Fig.~\ref{fig:snn} under the generalized $W_2$ distance metric under Assumption~\ref{assumptions_w2}. 
Theorem~\ref{theorem1} generalizes the universal approximation theorem of SNNs for approximating a random field model with continuous random variables in \cite[Appendix H]{xia2025new} to mixed random variables.

Note that the outputs $\hat{\bm{y}}_{\bm{x}}$ of the SNN model in Fig.~\ref{fig:snn} are continuous. We use the continuous outputs $\hat{\bm{y}}_{\bm{x}}$ when training the SNN. 
When utilizing the SNN to make predictions for the categorical components $y_{d_1+1}, ..., y_{d}$ on the testing set, we can use:
\begin{equation}
    \tilde{\bm{y}}_{\bm{x}} = \big(\hat{y}_1(\bm{x};\hat{\omega}),...,\hat{y}_{d_1}(\bm{x};\hat{\omega}), \text{round}^*(\hat{y}_{d_1+1}(\bm{x};\hat{\omega})),...,\text{round}^*(\hat{y}_d(\bm{x};\hat{\omega}))\big).
    \label{rounding}
\end{equation}
 In Eq.~\eqref{rounding}, $\text{round}^*(y)\coloneqq \max\big(\min(l, \text{round}(y)), u\big)$, where $\text{round}(y)$ is the rounding function $\mathbb{R}\rightarrow\mathbb{Z}$ and $l, u$ are the uniform upper and lower bounds for the categorical components $y_{d_1+1},...,y_{d}$, respectively. Therefore, the last $d-d_1$ components of $\tilde{\bm{y}}_{\bm{x}}$ in Eq.~\eqref{rounding} are categorical.

\subsection{A generalized local squared $W_2$ distance loss function}

Given a finite number of observed data, we do not have direct access to the probability measures $f_{\bm{x}}$, $\hat{f}_{\bm{x}}$, or $\nu(\d\bm{x})$ in Eq.~\eqref{generalized_define}. Therefore, direct minimization of $\hat{W}_2^2(\bm{y}_{\bm{x}}, \hat{\bm{y}}_{\bm{x}})$ in Eq.~\eqref{generalized_define} to train the SNN in Fig.~\ref{fig:snn} is not feasible. However, we can consider minimizing a generalized ``local" squared $W_2$ loss function, which is similar to the local squared $W_2$ loss function in \cite{xia2024local}, to train the SNN model in Fig.~\ref{fig:snn}.
\begin{definition}
\rm
    \label{local_w2}
    \textbf{The generalized local squared} $\bm{W_2}$ \textbf{loss function} is defined as:
    \begin{equation}
      \hat{W}_{2, \delta}^{2, \text{e}}(\bm{y}_{\bm{x}}, \hat{\bm{y}}_{\bm{x}})\coloneqq\int_D \hat{W}_2^2(f_{\bm{x}, \delta}^{\text{e}}, \hat{f}_{\bm{x}, \delta}^{\text{e}})\nu^{\text{e}}(\text{d}\bm{x}).
    \label{localw2}
\end{equation}
In Eq.~\eqref{localw2}, $\nu^{\text{e}}(\cdot)$ is the distribution and the empirical distribution of $\bm{x}$. $f_{\bm{x}, \delta}^{\text{e}}, \hat{f}_{\bm{x}, \delta}^{\text{e}}$ are the ``local" empirical probability measures of $\bm{y}(\tilde{\bm{x}};\omega)$ and $\hat{\bm{y}}(\tilde{\bm{x}};\hat{\omega})$ conditioned on $\|\tilde{\bm{x}}-\bm{x}\|_2\leq\delta$, respectively.
\end{definition}

We can prove the following generalization error bound on using the generalized local squared $W_2$ loss function Eq.~\eqref{localw2} with a finite number of training data, which is similar to \cite[Theorem 4.3]{xia2024local}.
\begin{theorem}
\rm
    \label{theorem_2}
    For each $\bm{x}\in D$, we denote the number of samples $(\tilde{\bm{x}}, \bm{y}_{\bm{\tilde{x}}})\in S$ such that $\|\tilde{\bm{x}}-\bm{x}\|_2\leq\delta$ to be $N(\bm{x}, \delta)$. We denote the total number of samples of the empirical distribution to be $N$. Assuming that each input $\bm{x}$ is independently sampled from the probability distribution $\nu$, then we have the following error bound
    \begin{equation}
    \E\Big[\big|\hat{W}_2^2(\bm{y}_{\bm{x}}, \hat{\bm{y}}_{\bm{x}}) - \hat{W}_{2, \delta}^{2, \text{e}}(\bm{y}_{\bm{x}}, \hat{\bm{y}}_{\bm{x}})\big|\Big] \leq \frac{4M}{\sqrt{N}} + 8CKM\E\big[h(N(\bm{x}, \delta), {d})\big] + 8\sqrt{M}L\delta
        \label{theorem2_result}
    \end{equation}
    where $\hat{W}_{2, \delta}^{2, \text{e}}(\bm{y}_{\bm{x}}, \hat{\bm{y}}_{\bm{x}})$ is the generalized local squared $W_2$ loss function defined in Eq.~\eqref{localw2}, and $\hat{W}_{2}^{2}(\bm{y}_{\bm{x}}, \hat{\bm{y}}_{\bm{x}})$ is the squared generalized $W_2$ distance between the two random fields $\bm{y}_{\bm{x}}$ and $\hat{\bm{y}}_{\bm{x}}$ defined in Eq.~\eqref{generalized_define}.
    $M$ is the upper bound for $\bm{y}_{\bm{x}}$ and $\hat{\bm{y}}_{\bm{x}}$ in Eq.~\eqref{upperboundy}, $C$ is a constant, $N$ is the total number of data points $(\bm{x}, \bm{y}_{\bm{x}})$, $K$ is the constant in Eq.~\eqref{upper_lower_bounds}, and $L$ is the Lipschitz constant in Eq.~\eqref{l_condition}. In Eq.~\eqref{theorem2_result}, 
    \begin{equation}
h(N, d)\coloneqq\left\{
\begin{aligned}
&2N^{-\frac{1}{4}}\log(1+N)^{\frac{1}{2}}, d\leq4,\\
&2N^{-\frac{1}{d}}, d> 4.
\end{aligned}
\right.
\label{t_def}
\end{equation}
\end{theorem}

The proof of Theorem~\ref{theorem_2} is similar to the proof of \cite[Theorem 4.3]{xia2024local} and is given in Appendix~\ref{appendixB}. Theorem~\ref{theorem_2} provides a generalization error bound on training the SNN with a finite number of data points, which greatly generalizes Theorem 1 in \cite{xia2024local} for continuous random variables to scenarios in which
$\bm{y}_{\bm{x}}$ in Eq.~\eqref{random_field_model} is a mixed random variable.

\subsection{A differentiable surrogate of the generalized local squared $W_2$ loss Eq.~\eqref{localw2}}

In Eq.~\eqref{random_field_model}, the last $d-d_1$ components of $\bm{y}_{\bm{x}}$ are discrete. However, the outputs $\hat{\bm{y}}_{\bm{x}}$ of the SNN are continuous. Additionally, $\hat{\delta}_{y_j, 0}$ in Eq.~\eqref{delta_function} is not differentiable, and $\partial_{y_j}\hat{\delta}_{y_j, 0}=0$ when $|y_j|>1$. Therefore, we need to create a differentiable surrogate of the generalized local squared $W_2$ loss in Eq.~\eqref{localw2} for training the SNN. For the ground truth $\bm{y}=(y_1,...,y_d)$ where $y_i\in\mathbb{R}, i=1,..,d_1$, $y_j\in\mathbb{Z}, j=d_1+1,...,d$ and the SNN's predicted $\hat{\bm{y}}=(\hat{y}_1,...,\hat{y}_d)\in\mathbb{R}^d$, we define the following pseudonorm:
\begin{equation}
    |\bm{y}-\hat{\bm{y}}|_1\coloneqq \lambda \sum_{i=1}^m (y_i-\hat{y}_i)^2 + \sum_{j=1}^n \tilde{\delta}_{y_j, \hat{y}_j},
    \label{dist1}
\end{equation}
where 
\begin{equation}
    \tilde{\delta}_{y_j, \hat{y}_j}=\begin{cases}
        1 - \frac{1}{2\pi}\cos\big(\frac{\pi}{2} + {2\pi} |y_j-\text{round}_1(\hat{y}_j)|\big),\,\, |y_j-\hat{y}_j|>\frac{1}{2},\\\
        4(y_j-\hat{y}_j)^2,\,\, |y_j- \hat{y}_j|\leq\frac{1}{2},
        \end{cases}
        \label{modified_dist}
\end{equation}
and
\begin{equation}
    \text{round}_1(\hat{y}_j)\coloneqq \hat{y}_j - (\hat{y}_j - \text{round}(\hat{y}_j)).\texttt{detach()}.
    \label{round_def}
\end{equation}
When both $y_j,\hat{y}_j\in\mathbb{Z}$ for $j=d_1+1,...,d$,
$\tilde{\delta}_{y_j, \hat{y}_j} = \hat{\delta}_{y_j- \hat{y}_j, 0}$ in Eq.~\eqref{delta_function} and $|\bm{y}-\hat{\bm{y}}|_1=\|\bm{y}-\hat{\bm{y}}\|$.
In Eq.~\eqref{round_def}, $(\hat{y}_j - \text{round}(\hat{y}_j)).\texttt{detach()}$ indicates \textbf{not} propagating the gradient of the tensor $(\hat{y}_j - \text{round}(\hat{y}_j))$ in \texttt{pytorch}. The distance Eq.~\eqref{dist1} is always differentiable w.r.t. $\hat{y}_j$ when the ground truth $y_j$ is categorical and the SNN's output $\hat{y}_j$ is continuous for $j=d_1+1,...,d$:
\begin{equation}
    \partial_{\hat{y}_j}\tilde{\delta}_{y_j, \hat{y}_j}=\begin{cases}
8(\hat{y}_j-y_j), \,\, |\hat{y}_j- y_j|\leq\frac{1}{2},\\
1,\,\, y_j>\hat{y}_j+\frac{1}{2},\\
-1,\,\, y_j<\hat{y}_j-\frac{1}{2}.
    \end{cases}
\end{equation}
$|\cdot|_1$ is used in replacement of the norm $\|\cdot\|$ defined in Eq.~\eqref{w2_distance} when numerically evaluating the generalized $W_2$ distance $\hat{W}_2(f_{\bm{x}}, \hat{f}_{\bm{x}})$ in Eq.~\eqref{pidef} and the generalized local squared $W_2$ loss function in Eq.~\eqref{localw2} in \texttt{Pytorch} to ensure differentiability of the loss function.

\section{Numerical examples}
\label{numerical_results}
In this section, we conduct numerical experiments to test our proposed generalized $W_2$ method. To boost efficiency, given $N$ observed data $\{(\bm{x}_i, \bm{y}_i)\}_{i=1}^N$, instead of using the generalized local squared $W_2$ loss function Eq.~\eqref{localw2}, we adopt a minibatch technique and adopt the following revised loss function:
\begin{equation}
          \frac{1}{n}\sum_{\bm{x}\in X_0} W_2^2(f_{\bm{x}, \delta}^{\text{e}}, \hat{f}_{\bm{x}, \delta}^{\text{e}}),
          \label{updated_loss}
\end{equation}
where $X_0\subseteq X\coloneqq\{\bm{x}_i\}_{i=1}^N$ is randomly chosen, $f_{\bm{x}, \delta}^{\text{e}}, \hat{f}_{\bm{x}, \delta}^{\text{e}}$ are the empirical probability measures of $\bm{y}(\tilde{\bm{x}};\omega)$ and $\hat{\bm{y}}(\tilde{\bm{x}};\hat{\omega})$ conditioned on $\|\tilde{\bm{x}}-\bm{x}\|_2\leq\delta$, 
and $n\coloneqq|X_0|$ is the cardinality of $X_0$.
$X_0$ is renewed and randomly selected again after every fixed number of training epochs. Numerical experiments in
Examples \ref{example1}, \ref{example3}, \ref{example4} are conducted using Python 3.11 on a desktop with a 32-core Intel®
i9-13900KF CPU.
Numerical experiments in Example \ref{example2} are carried out using Python 3.11 on NYU HPC with a GPU \cite{NYUHPC}. Training settings and hyperparameters for each example are listed in Table~\ref{tab:setting}. A pseudocode of our generalized $W_2$ approach to train the SNN in Fig.~\ref{fig:snn} by minimizing the loss function Eq.~\eqref{updated_loss} is given in Algorithm~\ref{algorithm_1}.

\begin{algorithm}
\footnotesize
\caption{\footnotesize The pseudocode of our generalized $W_2$ approach to train an SNN.}
\begin{algorithmic}
  \STATE Given $N$ observed data $\{(\bm{x}_i, \bm{y}_i), i=1,..., N\}$, the stopping criteria
    $\epsilon>0$, the size of the neighborhood $\delta$, the size of a minibatch $n$, the number of epochs for updating a minibatch $\text{epoch}_{\text{update}}$,  and the maximal epochs $\text{epoch}_{\max}$.
    \STATE Initialize the SNN in Fig.~\ref{fig:snn}.
\STATE For each $\bm{x}_i$, find samples in its neighborhood $B_i\coloneqq \{\bm{x}_j: \|\bm{x}_j-\bm{x}_i\|_2\leq \delta\}$.
    \STATE Input $\{\bm{x}_i\}, i=1,...,N$ into the neural network model to obtain predictions $\{\hat{\bm{y}}_i\}, i=1,...,N$.
  \FOR{$j=0,1,...,\text{epoch}_{\max}-1$}
  \IF{$j~\%~\text{epoch}_{\text{update}}$ == 0}
  \STATE Randomly choose $n$ samples from $\{(\bm{x}_i, \bm{y}_i), i=1,..., N\}$ to get a new $X_0$ in Eq.~\eqref{updated_loss}
  \ENDIF
  \STATE Calculate the loss function Eq.~\eqref{updated_loss} 
  \STATE Perform gradient descent to minimize the loss function
     and update the parameters (biases \& means and variances of weights) in the SNN.
\STATE Resample the weights in the SNN using the updated means and variances of weights.
    \STATE Input $\{\bm{x}_i\}, i=1,...,N$ into the updated SNN to obtain predictions $\{\hat{\bm{y}}_i\}, i=1,...,N$.  (for each $\bm{x}_i$, the weights in the SNN are sampled independently)
    \ENDFOR
  \RETURN The trained SNN
\end{algorithmic}
\label{algorithm_1}
\end{algorithm}

First, we present an example where the target random variable $\bm{y}_{\bm{x}}$ in Eq.~\eqref{random_field_model} is a univariate categorical variable.

\begin{example}
    \rm
    \label{example1}

In this example, we consider a classification problem:
\begin{equation}
        y_x = \begin{cases}
        A\big[\lfloor 4x+\xi\rfloor + 1\big], 0\leq \lfloor 4x+\xi\rfloor<5, \,\, A=\{3, 4, 1, 2, 0\}\\
        5, \quad\text{otherwise},
    \end{cases} 
    \label{example1_model}
\end{equation}
where $x\sim \mathcal{U}(-0.1, 1.1)$ and $\xi\sim\mathcal{N}(0,\sigma^2)$ is a random variable, $A[i]$ refers to the $i^{\text{th}}$ element of the set $A$, and $\lfloor\cdot\rfloor$ is the floor function. Given a set of training data points, we use the SNN model in Fig.~\ref{fig:snn}, trained by minimizing Eq.~\eqref{updated_loss}, as the approximate random field model Eq.~\eqref{approximate_random_field} to reconstruct Eq.~\eqref{example1_model} (shown in Algorithm~\ref{algorithm_1}).

To evaluate the accuracy of the reconstruction of the random field model Eq.~\eqref{example1_model} across different methods, we independently generate $\{y^j_{x_i}\}_{j=1}^{100}$ from Eq.~\eqref{example1_model} on each $x_i\in X\coloneqq\{0.01i-0.1, i=0,...,119\}$. Then, we evaluate the trained SNN 100 times independently on each $x_i\in X$ to get 100 $\{\hat{y}_{x_i}^j\}_{j=1}^{100}$. At each $x_i\in X$, we perform a permutation Chi-square test \cite{pesarin2010permutation} to test if $\{y^j_{x_i}\}_{j=1}^{100}$ and $\{\hat{y}_{x_i}^j\}_{j=1}^{100}$ follow the same distribution. We record the $p$-value of the permutation test, denoted by $p_{x_i}$. 
For those $p_{x_i}$ smaller than 0.05, we reject the null hypothesis that $\{y^j_{x_i}\}_{j=1}^{100}$ and $\{\hat{y}_{x_i}^j\}_{j=1}^{100}$ are drawn from the same distribution. Then, we evaluate the $p$-value test rejection rate (the number of $x_i$ satisfying $p_{x_i}<0.05$ divided by 120).
The lower the rejection rate is, the better the reconstruction of the random field model $y_x$ in Eq.~\eqref{example1_model} is.  
We test: i) how the value of $\sigma$, the uncertainty level in the target $y_x$, affects the reconstruction accuracy of the random field model Eq.~\eqref{example1_model} and ii) how the number of training data points affects the reconstruction accuracy of Eq.~\eqref{example1_model}. Additionally, we benchmark our proposed generalized $W_2$ method against other methods, including the mixture density network method trained by minimizing a cross-entropy loss function \cite{ghahramani1993supervised}, the ensemble entropy method that uses the ensemble of five independently trained mixture density networks \cite{lakshminarayanan2017simple}, the evidential learning method \cite{sensoy2018evidential}, the Bayesian neural network (BNN) method \cite{mullachery2018bayesian}, and the local squared $W_2$ method \cite{xia2024local}. 

    \begin{figure}
    \centering
\includegraphics[width=0.8\linewidth]{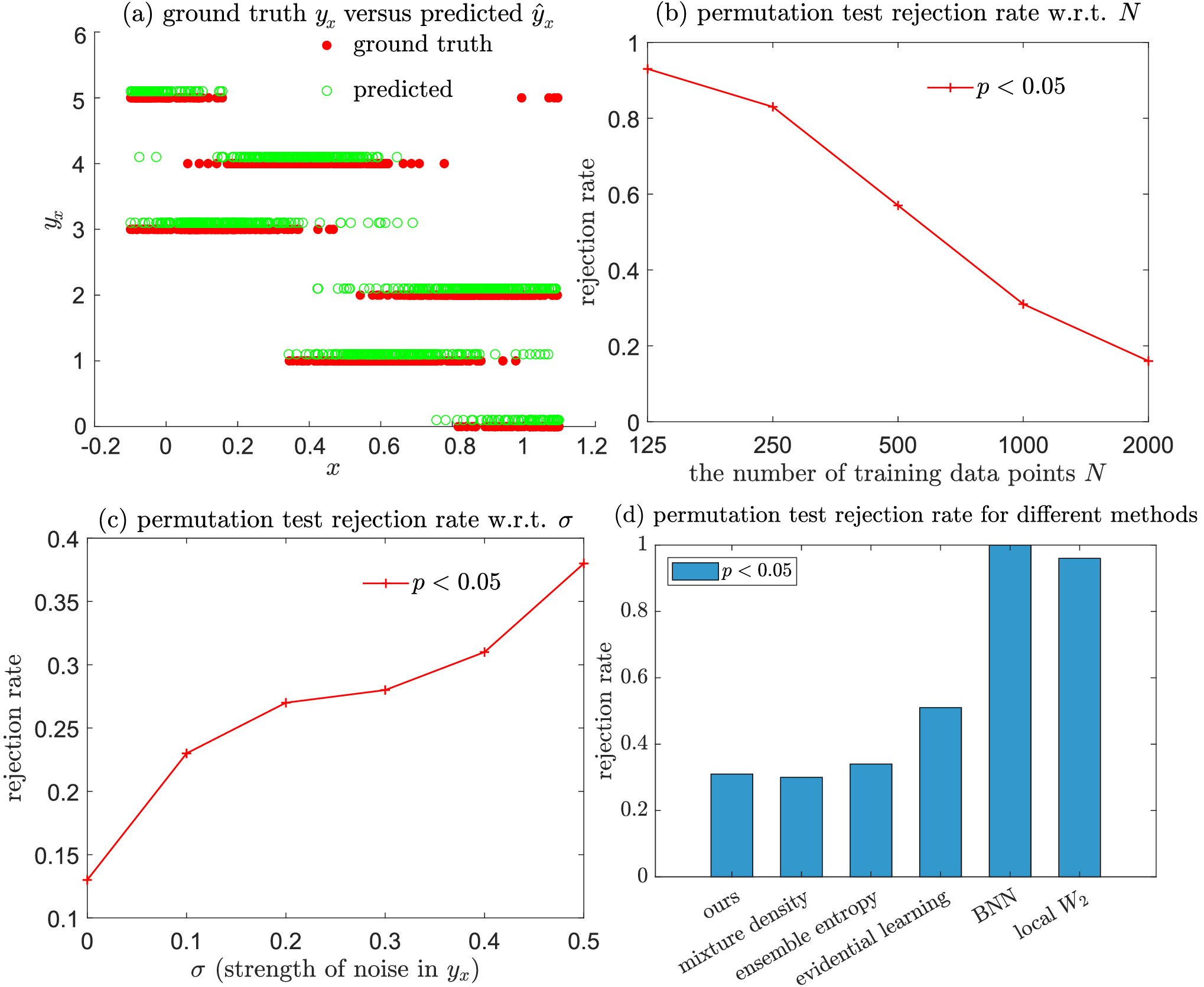}
    \caption{(a) ground truth $y_x$ versus $\hat{y}_x$ generated by the trained SNN (for visualization clarity, we scatter $(x, y_x)$ and $(x, \hat{y}_x+0.1))$. The number of training data is $N=1000$. (b) the permutation test null hypothesis rejection rate w.r.t. the number of training data points. In (a)(b), $\sigma=0.4$ in Eq.~\eqref{example1_model}. (c) the permutation rejection rate w.r.t. the uncertainty level $\sigma$ in Eq.~\eqref{example1_model} (the number of training data $N=1000$). (d) the permutation test rejection rate of different methods ($\sigma=0.4$ and $N=1000$).}
    \label{fig:example1}
\end{figure}

From Fig.~\ref{fig:example1} (a), the distribution of $\hat{y}_x$ obtained from the trained SNN matches well with the distribution of the ground truth $y_x$ in Eq.~\eqref{example1_model}. As the number of training data points increases, the reconstructed random field model becomes more accurate, as shown in Fig.~\ref{fig:example1} (b). From Fig.~\ref{fig:example1} (c), when $\sigma$ increases, the reconstruction of the uncertainty model Eq.~\eqref{example1_model} becomes less accurate. As shown in Fig.~\ref{fig:example1} (d), our proposed generalized $W_2$ method gives comparable performance to the mixture density network method and the ensemble entropy method, and it outperforms the evidential learning method, the BNN method. Specifically, the previous local squared $W_2$ method in \cite{xia2024local} relies on the $\ell^2$ norm for continuous variables and performs poorly on reconstructing the distribution of the categorical $y_x$ in Eq.~\eqref{example1_model}.

As an additional experiment, we investigate how the structure of the neural network affects the reconstruction of the model Eq.~\eqref{example1_model}. We find that an SNN with five hidden layers, 32 neurons in each layer, equipped with the GELU activation function and the ResNet technique, can most accurately reconstruct the uncertainty model Eq.~\eqref{example1_model}. Furthermore, we explore whether replacing the coefficient 4 in Eq.~\eqref{w2_distance} with other constants could impact the reconstruction accuracy of the model Eq.~\eqref{example1_model}. Our results indicate that using the coefficient $4$ to ensure that $\|\cdot\|$ defined in Eq.~\eqref{w2_distance} is a norm leads to the most accurate reconstruction of Eq.~\eqref{example1_model}. 
Detailed results of these additional sensitivity tests are in Appendix~\ref{example1_appendix}.
\end{example}

Next, we investigate how the dimensionality of the categorical random variable influences the accuracy of reconstructing its distribution. 

\begin{example}
    \rm
    \label{example2}
We consider an example in which the target random variable $\bm{y}_{\bm{x}}$ in Eq.~\eqref{random_field_model} is multidimensional categorical. We use the \texttt{make\_multilabel\_classification} function in \texttt{sklearn} to generate a synthetic data set, consisting of 4000 training data points and another 1000 testing data points. The features $\bm{x}$ are continuous, while all components in the target variable $\bm{y}_{\bm{x}}$ are binary. The input $\bm{x}$ is 8-dimensional. On average, two components of $\bm{y}_{\bm{x}}$ are 1 while the rest components are 0.

When $\bm{y}_{\bm{x}}=(y_1,...,y_d)$ is a multivariate categorical random variable whose components are binary, we can transform it into a 1D categorical variable: 
\begin{equation}
    \tilde{y}_{\bm{x}}\coloneqq \sum_{i=1}^d 2^{i-1}y_i.
    \label{hat_y_uni}
\end{equation}
There is a one-to-one mapping from $\bm{y}_{\bm{x}}$ to $\tilde{y}_{\bm{x}}$ in Eq.~\eqref{hat_y_uni}.

For predicting the categorical sex variable on the testing set, we independently input the features $\bm{x}$ into the trained SNN, repeating 50 times. Then, we choose the category that appears the most often as the prediction of the testing data (if there are two or more categories that appear most frequently, the class that appears first in the 50 repeated predictions will be assigned).

\begin{table}[h]
\centering
\caption{Classification accuracy ($\frac{\text{correct predictions}}{\text{total testing samples}}$), runtime, and memory usage when using the original $\bm{y}_{\bm{x}}$ or the transformed $\hat{y}_{\bm{x}}$ in Eq.~\eqref{hat_y_uni} as the target random variable. The number in the bracket indicates the total number of potential categories of the target.}
\label{tab:example2}
\begin{tabular}{|c|c|c|c|c|c|c|} 
\hline
  &\multicolumn{2}{c|}{Accuracy} & \multicolumn{2}{c|}{{Memory Usage} (Mb)} & \multicolumn{2}{c|}{{Runtime} (h)} \\ 
\hline
 Dimensionality of $\bm{y}_{\bm{x}}$ & $\tilde{y}_{\bm{x}}$ & $\bm{y}_{\bm{x}}$ &$\tilde{y}_{\bm{x}}$ & $\bm{y}_{\bm{x}}$&$\tilde{y}_{\bm{x}}$ & $\bm{y}_{\bm{x}}$\\
\hline
3 ($2^3$)& 0.80 & 0.54 & 5300 & 2884 & 2.47 & 6.26\\
\hline
4 ($2^4$)& 0.52 & 0.25 &  6058&  4069 &2.52 &  8.55 \\
\hline
5 ($2^5$)& 0.37 & 0.22 & 3665 & 2867 &2.53 & 10.67\\
\hline
6 ($2^6$)& 0.26 & 0 & 3838 & 4121 &1.75 &12.22 \\
\hline
7 ($2^7$)& 0.25 & 0 & 5587 & 2906 & 2.63& 13.28\\
\hline
\end{tabular}
\end{table}

From Table~\ref{tab:example2}, the prediction accuracy decreases as the dimensionality of the output increases no matter whether the original $\bm{y}_{\bm{x}}$ or the transformed $\tilde{y}_{\bm{x}}$ in Eq.~\eqref{hat_y_uni} are used as the target. Converting $\bm{y}_{\bm{x}}$ to the 1D $\hat{y}_{\bm{x}}$ in Eq.~\eqref{hat_y_uni} leads to improved reconstruction accuracy. The underlying reason could be that the convergence rate of the empirical probability measure $f^{\text{e}}_{\bm{x}}$ to the ground truth probability measure $f_{\bm{x}}$ becomes slower as the dimensionality of $\bm{y}_{\bm{x}}$ increases w.r.t. the number of training data points, as proved in Theorem~\ref{theorem_2}. 

Additionally, the runtime of using the 1D $\tilde{y}_{\bm{x}}$ is significantly smaller than using $\bm{y}_{\bm{x}}$. Thus, it could be beneficial to convert a multivariate categorical random variable into a univariate categorical random variable through a transformation as Eq.~\eqref{hat_y_uni} for more efficient reconstruction of the random field model.
\end{example}

Next, we consider an example in which $\bm{y}_{\bm{x}}$ in the random field model Eq.~\eqref{random_field_model} is a mixed random variable for every $\bm{x}$.

\begin{example}
    \rm
    \label{example3}
We study the problem of abalone sex classification and age prediction in \cite{abalone1}. Seven continuous variables are recorded as measurements: length (mm), diameter (mm), height (mm), whole weight (gram), shucked weight (gram), viscera (gram), weight (gram), and shell weight (gram). We predict a continuous variable ``rings" (rings$+1.5=$age) and a categorical variable ``sex" of the abalone (male, female, and infant). As stated in \cite{abalone1}, the features recorded are not sufficient to predict the target variables, and other unrecorded factors, such as weather patterns and food availability, may be required to characterize sex and rings. Therefore, we model the dependence of rings and sex based on the seven observed continuous variables using the random field model Eq.~\eqref{random_field_model}, where $\bm{x}$ is the seven observed variables and $\bm{y}_{\bm{x}} = (y_1(\bm{x};\omega), y_2(\bm{x};\omega))$ consists of a continuous component $y_1$ characterizing the continuous variable rings and a categorical component $y_2$ representing sex ($\omega$ is the set of factors that are not recorded).

When the neighborhood size $\delta=0$ in Eq.~\eqref{updated_loss}, our proposed loss function degenerates to the mean square error loss given finite observed data when $\bm{x}_i\neq \bm{x}_j, i\neq j$. However, using the mean square error is insufficient to quantify the uncertainty of $\bm{y}_{\bm{x}}$ \cite{xia2024local}. On the other hand, when the neighborhood size $\delta=\infty$ in Eq.~\eqref{updated_loss}, the dependence on $\bm{x}$ is ignored, which leads to systematic errors as was shown in \cite{xia2024local}. Therefore, we explore how the choice of $\delta$ influences the ability of the trained SNN to reconstruct the distribution of $\bm{y}_{\bm{x}}$ for every $\bm{x}$.

We randomly split the whole dataset into a training set (80\% of the total data) to train the SNN and a testing set (the rest 20\% of the total data). The features are normalized to have a mean of 0 and a variance of 1. On the testing set, for the continuous $y_1(\bm{x};\omega)$, the $R^2$ statistic represents the proportion of variance in the dependent variable that is explained by the independent variables in the model:
\begin{equation}
    R^2 = 1 - \frac{\int_D (y_1(\bm{x};\omega)-\E[\hat{y}_1(\bm{x};\hat{\omega})])^2\nu^{\text{e}}(\d\bm{x})}{\int_D \big(y_1(\bm{x};\omega)-\bar{y}_1\big)^2\nu^{\text{e}}(\d\bm{x})}.
    \label{r2square}
\end{equation}
where $\E[\hat{y}_1(\bm{x};\hat{\omega})]$ is the expectation of the SNN's prediction for the continuous variable rings at $\bm{x}$ and $\bar{y}_1$ denote the average value of $y_1(\bm{x};\omega)$ on the testing set. With the trained SNN, we also calculate a scaled  predicted variance:
\begin{equation}
   \text{Var}_{\hat{y}_1}\coloneqq \frac{\int_D \text{Var}[\hat{y}_1(\bm{x};\hat{\omega})]\nu^{\text{e}}(\d\bm{x})}{{\int_D \big(y_1(\bm{x};\omega)-\bar{y}_1\big)^2\nu^{\text{e}}(\d\bm{x})}}
    \label{var_approx}
\end{equation}
on the testing set. In Eq.~\ref{var_approx}, $\text{Var}[\hat{y}_1(\bm{x};\hat{\omega})]$ indicates the variance of the SNN's prediction for the continuous variable rings $\hat{y}_1(\bm{x};\hat{\omega})$ at $\bm{x}$. Then, we compare Eq.~\eqref{var_approx} with Eq.~\eqref{r2square} to evaluate how the trained SNN model can quantify the uncertainty in the target variable $y_1(\bm{x};\omega)$ that cannot be explained by the average value of the prediction $\E[\hat{y}_1(\bm{x};\hat{\omega})]$. As a baseline model for comparison, we train a hybrid deterministic neural network whose output layer integrates the output layer of a mixture neural network for predicting the categorical $y_2$ and the output layer of a feedforward neural network for predicting the continuous $y_1$. The internal structure of the hybrid deterministic neural network is the same as the SNN (\textit{i.e.} the hybrid neural network has the same number of hidden layers and neurons in each layer, but the weights in the deterministic neural network are deterministic). The hybrid deterministic neural network is trained by minimizing a hybrid loss function, which is the summation of the MSE for the prediction of the continuous variable $\hat{y}_1$ and the cross-entropy loss for the prediction of the categorical variable $\hat{y}_2$.

    \begin{figure}
    \centering
\includegraphics[width=0.8\linewidth]{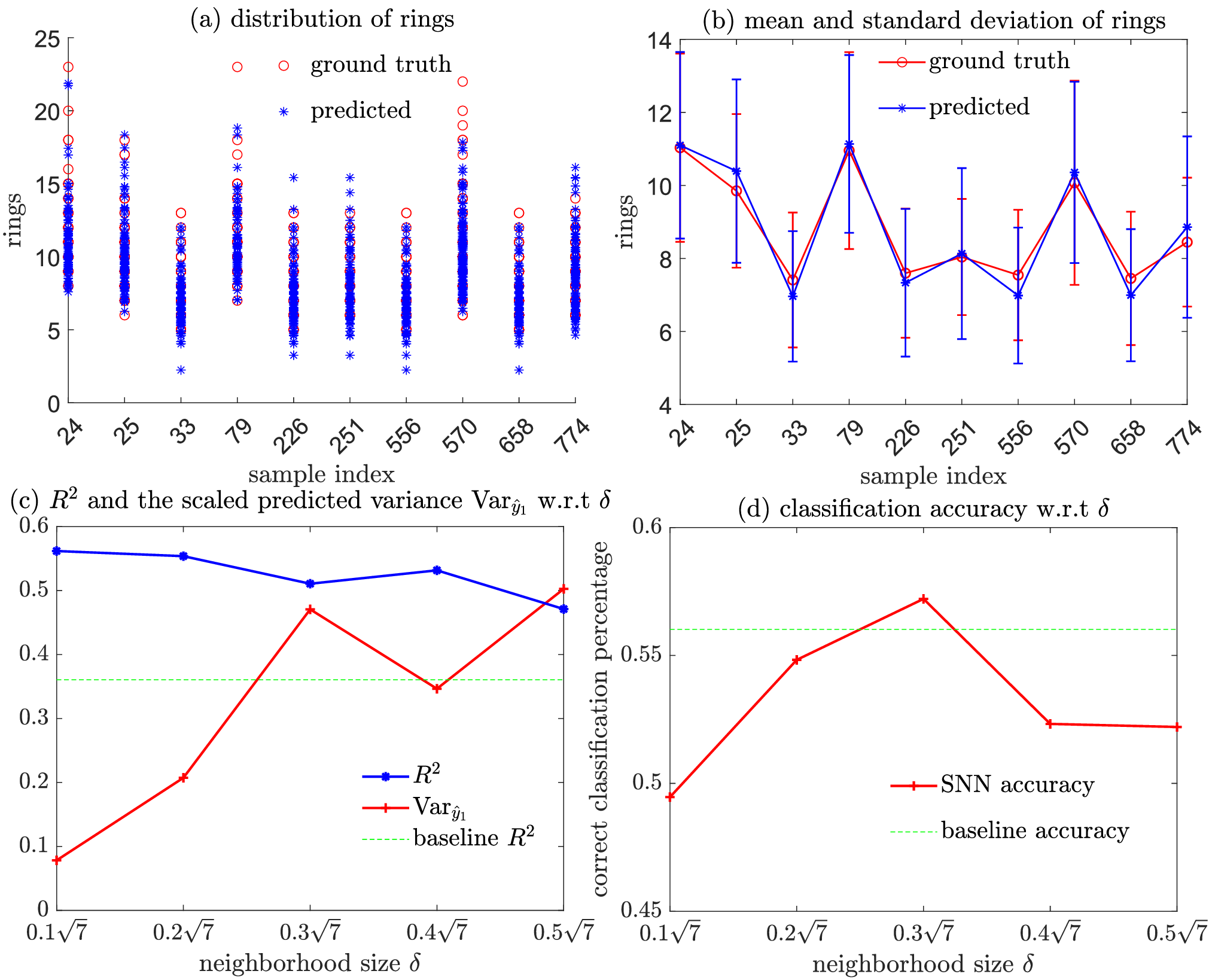}
    \caption{(a) the ground truth and predicted values of the continuous variable rings ($y_1$ and $\hat{y}_1$) in the neighborhoods of ten randomly chosen samples which have no fewer than 50 neighbors in the neighborhood $\|\bm{x}-\tilde{\bm{x}}\|_2\leq\delta$ in the testing set. (b) the mean and standard deviations of the ground truth and predicted values of the continuous variable rings ($y_1$ and $\hat{y}_1$) in the neighborhoods of ten randomly chosen samples which have no fewer than 50 neighbors in the neighborhood $\|\bm{x}-\tilde{\bm{x}}\|_2\leq\delta$ in the testing set. In (a)(b), we set $\delta=0.3\sqrt{7}$ in the loss function Eq.~\eqref{updated_loss}. (c) the $R^2$ score in Eq.~\eqref{r2square} as well as the scaled predicted variance in Eq.~\eqref{var_approx} for the predictions from SNNs trained by minimizing the loss function Eq.~\eqref{updated_loss} with different values of $\delta$. The baseline $R^2$ score from the deterministic neural network is shown in green. (d) the classification accuracy for predicting the categorical variable sex on the testing set for SNNs trained by minimizing Eq.~\eqref{updated_loss} with different values of $\delta$. The baseline classification accuracy from the deterministic neural network is shown in green. For classification, the SNN is evaluated 50 times independently on each data point of the testing set, and we take the class that occurs the most as the prediction.}
    \label{fig:example3}
\end{figure}

From Fig.~\ref{fig:example3} (a)(b), the distribution of the continuous variable $y_1(\tilde{\bm{x}};\omega)$ whose $\tilde{\bm{x}}$ are in the neighborhoods $(\|\bm{x}-\tilde{\bm{x}}\|_2\leq\delta)$ of ten samples in the testing set can be well matched by the distribution of the predicted $\hat{y}_1(\tilde{\bm{x}};\hat{\omega})$. In Fig.~\ref{fig:example3} (c), as $\delta$ in the loss function Eq.~\eqref{updated_loss} increases, the variance in the predictions from the SNN increases. Similar to the results in \cite[Example 2]{xia2024local}, a too-small $\delta$ prevents the SNN from quantifying the uncertainty in the output. This is because when $\delta\rightarrow0^+$, there are fewer samples in the neighborhood of each $\bm{x}$, making it harder to quantify the uncertainty of $y_1(\bm{x};\omega)$ for every $\bm{x}$. On the other hand, a too-large $\delta$ leads to systematic errors and compromised classification accuracy. Compared to the baseline hybrid neural network model, all SNNs have a larger $R^2$ score, indicating a better prediction given the seven observed variables. When $\delta=0.3\sqrt{7}$, the variance in the prediction Eq.~\eqref{fig:example3} approximately matches the $R^2$ score, indicating that the trained SNN could quantify the uncertainty in $y_1(\bm{x};\omega)$ well.
Finally, as shown in Fig.~\ref{fig:example3} (d), the classification accuracy of the SNN is comparable to that of the baseline hybrid neural network when $\delta=0.3\sqrt{7}$.
The result is similar to that of Example~\ref{example1}, showing that for reconstructing the distribution of categorical random variables, our SNN performs similarly to the mixture neural network. However, our SNN gives much better prediction on the distribution of the continuous component $y_1(\bm{x};\omega)$ than the hybrid deterministic neural network, indicating that the SNN trained by minimizing our loss Eq.~\eqref{updated_loss} can better reconstruct the distribution of the mixed random variable $\bm{y}(\bm{x};\omega)$ for different $\bm{x}$. 
\end{example}

Finally, we consider a real-world application of reconstructing a dynamical system in which Markov jump processes are coupled with ODEs to describe gene regulatory dynamics.

\begin{example}
    \rm
    \label{example4}

The interactions between multiple genes are often described by a dynamical system, in which continuous gene expression levels (the number of mRNA, protein, etc) and categorical gene states are mutually regulated by each other, with wide applications such as predicting cell fates \cite{jaruszewicz2013toggle, tian2006stochastic}. 
In \cite{jaruszewicz2013toggle}, a gene toggle model is studied to describe interactions between two mutually regulated genes, and it is found that intrinsic noise resulting from gene state switch could lead to heterogeneous cell fates. A Markov process describing the state change of two genes is coupled with an ODE describing the dynamics of scaled mRNA, protein, and protein dimer counts to describe the dynamics of two genes that suppress each other:
\begin{equation}
\begin{aligned}
\frac{\d m_i(t)}{\d t} &= k_8 g_i(t) + \frac{k_4}{M_0}\big(1 - g_i(t)\big) - k_8 m_i(t),  \\
\frac{\d p_i(t)}{\d t} &= 2 \theta_i k_6 P_0 \left( d_i(t) - p_i^2(t) \right) + k_9 (m_i(t) - p_i(t)), \\
\frac{\d d_i(t)}{\d t} &= \theta_i k_7 \left( p_i^2(t) - d_i(t) \right), \,\, i=1,2,
\end{aligned}
\label{gene_toggle_1}
\end{equation}
and
\begin{equation}
    P(g_i(t+\Delta t)=1|g_i(t) = 0) = \sigma_ik_1 \Delta t,\,\, P(g_i(t+\Delta t)=0|g_i(t) = 1) =  \sigma_i k_2 D_0 d_j(t) g_i(t)\Delta t.
    \label{gene_toggle_2}
\end{equation}
 In Eqs.~\eqref{gene_toggle_1} and \eqref{gene_toggle_2}, $g_i(t)\in\{0, 1\}, i=1, 2$ represent gene one and gene two's state. The scaled counts of mRNA, protein, and protein dimer, which will be treated as continuous variables, associated with gene 1 or gene 2 are defined as
\[
m_i(t) = \frac{M_i(t)}{M_0}, \quad 
p_i(t) = \frac{P_i(t)}{P_0}, \quad 
d_i(t) = \frac{D_i(t)}{D_0},
\]
where \( M_i(t), P_i(t) \), and \( D_i(t) \) are the number of mRNA, protein, and protein dimers at time $t$. The constants $M_0, P_0, D_0$ are defined as: 
\[
M_0 \coloneqq \frac{k_3}{k_8}, \quad 
P_0 \coloneqq \frac{k_3 k_5}{k_8 k_9}, \quad 
D_0 \coloneqq \frac{k_6 (k_3 k_5)^2}{k_7 (k_8 k_9)^2}.
\]
We superimpose a small noise to characterize cell heterogeneity onto the fixed initial conditions used in \cite{jaruszewicz2013toggle}
and set $m_i(0)=0.15(1+\xi_{\text{m}, i})$,  $p_i(0)=0.15(1+\xi_{\text{p}, i})$, and $d_i(0)=0.022(1+\xi_{\text{d}, i})$, where $\xi_{\text{m}, i}, \xi_{\text{p}, i}, \xi_{\text{d}, i}\sim\mathcal{U}(0, 0.05)$. For the two genes' initial states, we sample their initial states with the probability: $P(g_1(0)=0)=P(g_1(0)=1)=\frac{1}{2}$ and $P(g_2(0)=0)=P(g_2(0)=1)=\frac{1}{2}$ (note: in \cite{jaruszewicz2013toggle}, Eq.~\eqref{gene_toggle_2} is further approximated by an ODE). The biological interpretations and values of parameters used in Eqs.~\eqref{gene_toggle_1} and \eqref{gene_toggle_2} are the same as in \cite{jaruszewicz2013toggle} and are given in Table~\ref{tab:reaction_rates}.

\begin{table}[h]
\centering
\footnotesize
\caption{Biophysical meanings and values of parameters used in Eqs.~\eqref{gene_toggle_1} and \eqref{gene_toggle_2} (mlcl$=$molecule), which is the same as \cite[Table 1]{jaruszewicz2013toggle}.}
\label{tab:reaction_rates}
\begin{tabular}{llcc}
\toprule
\textbf{Parameter} & \textbf{Symbol} & \textbf{Default values} ($\sigma_i = 1$, $\theta_i = 1$)  \\
\midrule
Gene activation by protein dimer dissociation & $\sigma_1k_1$ & 0.003 (1/s)  \\
Gene repression by protein dimer binding & $\sigma_1k_2$ & 0.015 (1/(mlcl$\times$s))  \\
mRNA transcription from the active gene & $k_3$ & 0.02 (1/s)  \\
mRNA transcription from the repressed gene & $k_4$ & 0.0006 (1/s)  \\
Protein translation & $k_5$ & 0.01 (1/(mlcl$\times$s))  \\
Dimer formation & $\theta_1k_6$ & 0.0001 (1/(mlcl$\times$s))  \\
Dimer dissociation to monomers & $\theta_1k_7$ & 0.01 (1/s)  \\
mRNA degradation & $k_8$ & 0.005 (1/s)  \\
Protein monomer degradation & $k_9$ & 0.0005 (1/s)  \\
\bottomrule
\end{tabular}
\end{table}

Given a batch of trajectories $\{g_i(t), m_i(t), p_i(t), d_i(t)\}_{i=1}^2$ generated from numerically solving Eqs.~\eqref{gene_toggle_1} and \eqref{gene_toggle_2}, we  
reconstruct the dynamical systems Eqs.~\eqref{gene_toggle_1} and~\eqref{gene_toggle_2} using:
\begin{equation}
\begin{aligned}
       &\frac{\d \hat{\bm{y}}(t)}{\d t} = \text{NN}_1(\hat{\bm{y}}(t), \bm{\hat{g}}(t)),\\
       &\hat{\bm{g}}(t+\Delta t) = \hat{\bm{g}}(t) + \text{SNN}_2(\hat{\bm{y}}(t), \bm{\hat{g}}(t), \Delta t),
\end{aligned}
\label{approximate_ODE}
\end{equation}
where $\hat{\bm{y}}(t)\coloneqq (\hat{m}_1(t), \hat{p}_1(t),\hat{d}_1(t),\hat{m}_2(t), \hat{p}_2(t),\hat{d}_2(t))$ stands for the vector of approximate scaled mRNA, protein, and dimer counts of gene 1 and gene 2, and $\hat{\bm{g}}(t)$ denotes the predicted gene states of gene 1 and gene 2. $\text{NN}_1$ is a deterministic neural network with 3 hidden layers, 32 neurons in each later, and the RELU activation function, which approximates the RHS of the ODE system~\eqref{gene_toggle_1}; $\text{SNN}_2$ is an SNN in Fig.~\ref{fig:snn} to approximate the Markov jump process describing genes' state transitions Eq.~\eqref{gene_toggle_2}. The ODEs Eqs.~\eqref{gene_toggle_1} and the first equation in Eq.~\eqref{approximate_ODE} are numerically solved using the \texttt{odeint} function in the \texttt{torchdiffeq} package up to $t=1\text{min}$. To take into account the distributions of ground truth trajectories and predicted trajectories at different times, we use a time-averaged version of the loss function Eq.~\eqref{updated_loss}, which generalizes the local squared temporally decoupled squared $W_2$ loss in \cite{xia2025new}:
\begin{equation}
    \frac{1}{T}\frac{1}{n}\sum_{j=1}^T\sum_{\bm{y}_0\in X_0} \hat{W}_2^2\big(f_{\bm{y}_0, \delta}^{\text{e}}(t_j), \hat{f}_{\bm{y}_0, \delta}^{\text{e}}(t_j)\big).
    \label{temporal_loss}
\end{equation}
In Eq.~\eqref{temporal_loss}, $f_{\bm{x}, \delta}^{\text{e}}(t_j)$ and $\hat{f}_{\bm{x}, \delta}^{\text{e}}(t_j)$ are the local empirical measures of $\bm{y}(t_j)$ and $\hat{\bm{y}}(t_j)$ at time $t=t_j$ conditioned on the initial condition satisfying $\|\bm{y}(0)-\bm{y}_0\|\leq\delta$ and $\|\hat{\bm{y}}(0)-\bm{y}_0\|\leq\delta$, respectively.
In Eq.~\eqref{temporal_loss}, we use $t_j=j\Delta t, \Delta t=0.1, T=10$. For simplicity, we plot the ground truth and predicted trajectories of the mRNA dynamics associated with two genes (Fig.~\ref{fig:example4} (a)(b)), the ground truth and predicted rates of change $\frac{\d m_i(t)}{\d t}$ and $\frac{\d \hat{m}_i(t)}{\d t}$ w.r.t. the two types of mRNA (Fig.~\ref{fig:example4} (c)(d)), the ground truth and predicted proportion of cells with activated gene 1 and gene 2 (Fig.~\ref{fig:example4} (e)(f)), and the transition probability of gene 1 from the deactivated state to the deactivated state (Fig.~\ref{fig:example4} (g)) and from the activated state to the activated state (Fig.~\ref{fig:example4} (h)).

    \begin{figure}
    \centering
\includegraphics[width=\linewidth]{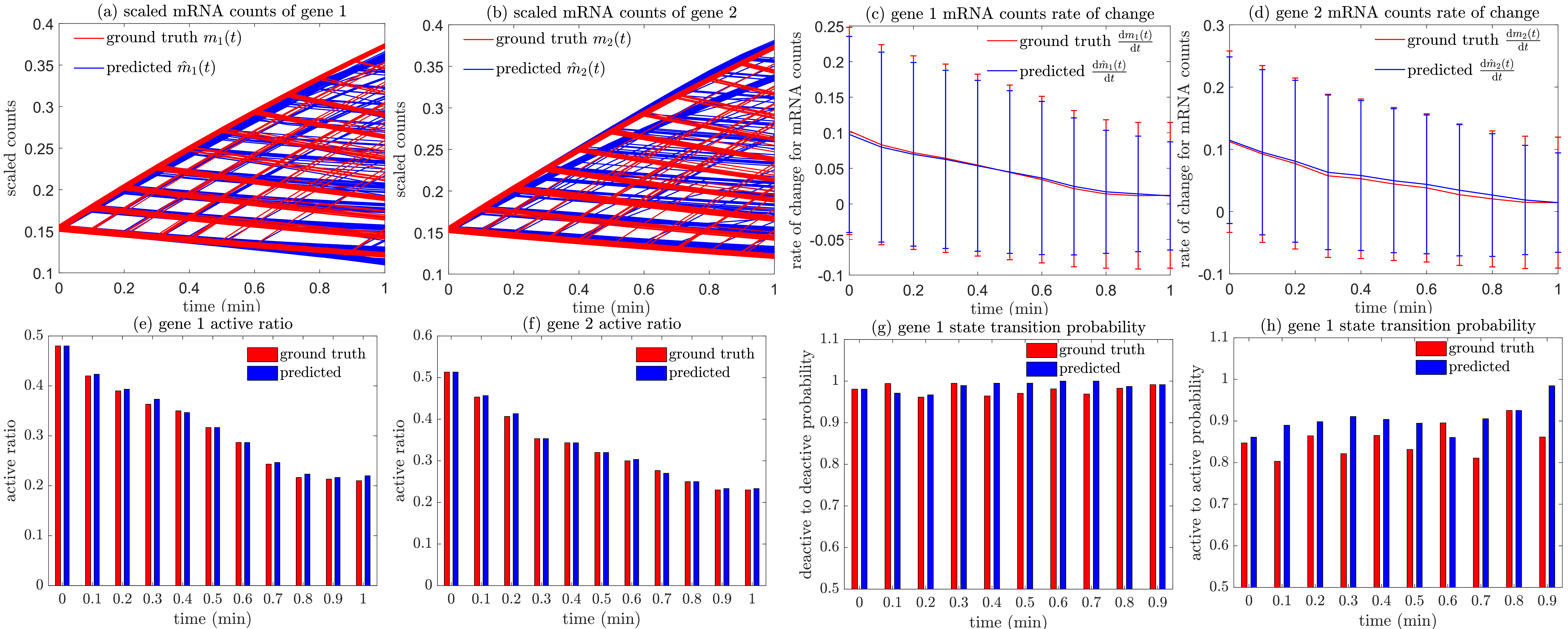}
    \caption{(a, b) the scaled mRNA counts transcripted from gene 1 and gene 2 over time, respectively. (c, d) mean and standard deviation of the rate of change for the scaled mRNA counts associated with gene 1 and gene 2 over time. (e, f) ground truth and predicted ratios of cells with activated gene 1 and/or activated gene 2, respectively. (g) ground truth versus predicted gene state transition probabilities of gene 1 from the deactivated state to the deactivated state at different times, evaluated on all predicted trajectories of gene expression dynamics. (h) ground truth and predicted gene state transition probabilities of gene 1 from the activated state to the activated state at different times, evaluated on all predicted trajectories of gene expression dynamics.}
    \label{fig:example4}
\end{figure}

 By minimizing the loss function Eq.~\eqref{temporal_loss}, both the deterministic neural network characterizing the dynamics of scaled mRNA, protein, and protein dimer counts and the SNN characterizing genes' state switching dynamics in Eq.~\eqref{approximate_ODE} can be trained to approximate the ground truth Eqs.~\eqref{gene_toggle_1} and \eqref{gene_toggle_2} well, respectively. From Fig.~\ref{fig:example4} (a)(b), the distribution of ground truth trajectories can be well matched by the distribution of predicted trajectories generated from Eq.~\eqref{approximate_ODE}; furthermore, the distribution of the rate of change in mRNA counts $\frac{\d m_i(t)}{\d t}$ can also be matched by the distribution of $\frac{\d \hat{m}_i(t)}{\d t}$ (shown in Fig.~\ref{fig:example4} (c)(d)). Furthermore, since the interacting gene 1 and gene 2 obey the same regulatory dynamics Eqs.~\eqref{gene_toggle_1} and Eqs.~\eqref{gene_toggle_2}, the empirical distribution of $m_1(t)$ is similar to that of $m_2(t)$, and the empirical distribution of $\frac{\d m_1(t)}{\d t}$ is close to that of $\frac{\d m_2(t)}{\d t}$,. The trained neural ODE model and SNN model Eq.~\eqref{approximate_ODE} also reproduce the symmetry in the two interacting genes' regulatory dynamics.
The ground truth proportion of cells with gene 1 or gene 2 activated can also be matched well by the predicted proportion of cells with the corresponding gene activated (Fig.~\ref{fig:example4} (e)(f)). This is because the learned Markov jump process (second equation in Eq.~\eqref{approximate_ODE}) has a similar transition probability for gene switching states to that of the ground-truth Markov jump process Eq.~\eqref{gene_toggle_2}, as shown in Fig.~\ref{fig:example4} (g)(h).
\end{example}

\section{Summary and Conclusion}
\label{conclusion}
In this work, we proposed a generalized $W_2$ method to train an SNN to reconstruct random field models of mixed random variables from a finite number of training data. Our proposed method was successfully applied to various UQ tasks such as classification, reconstructing the probability distribution of random variables consisting of both categorical and continuous components, and reconstructing a coupled system of ODEs and Markov jump processes characterizing gene regulatory dynamics.
For classification tasks, our method achieved performance comparable to that of prevailing machine-learning methods. For reconstructing the distribution of mixed random variables, our method yielded better performance compared to a benchmark neural network-based method. 

As a future direction, it is promising to explore how to incorporate constraints or prior knowledge of the random field model to be reconstructed. In addition, investigations on how the dimensionality of the mixed random variable affects the accuracy of the reconstruction of its distribution can be helpful. Further analysis and refinement of the distance metric in Eq.~\eqref{w2_distance} for mixed random variables would be beneficial. Reconstructing a stochastic differential equation with state transitions using our approach is also worth further investigation.
Finally, one may also analyze using the entropic regularized Wasserstein distances and applying the Sinkhorn algorithm \cite{cuturi2013sinkhorn} to solve corresponding optimal transport problems, which could lead to reduced computational complexity.

\section*{Acknowledgement}
The authors thank Prof. Alex Mogilner at New York University and Prof. Philip Maini at the University of Oxford
 for their valuable suggestions on this work.

\appendix
\section{Proof to Theorem~\ref{theorem1}}
\label{appendixA}
Here, we prove Theorem~\ref{theorem1}. For any $\bm{z}\in\mathbb{R}^d$, we denote $\bm{z}_1\coloneqq (z_1,...,z_{d_1})$ to be its first $d_1$ components and $\bm{z}_2=\coloneqq(z_{d_1+1},...,z_{n})$ to be its last $d_2$ components. Suppose the probability measure of $\bm{y}_{\bm{x}}=(\bm{y}_1,\bm{y}_2), \bm{y}\in\mathbb{R}^{d_1}, \bm{y}_2\in\mathbb{N}^{d-d_1}$ is $f_{\bm{x}}$ such that:
\begin{equation}
    \sum_{\bm{y}_2\in S_{d-d_1}}\int_{\mathbb{R}^{d_1}}f_{\bm{x}}(\bm{y}_1, \bm{y}_2)\d\bm{y}_1=1,
\end{equation}
where $S_{d-d_1}\subseteq \mathbb{R}^{d-d_1}$ is a bounded set including all possible outcomes of the categorical $\bm{y}_2$. 
First, consider the following convoluted probability measure:
\begin{equation}
    f_{\epsilon, \bm{x} }(\bm{z}) \coloneqq 
    \begin{cases}
        f_{\bm{x}}(\bm{y})\phi_{\epsilon}(\bm{y}-\bm{z}), |\bm{z}-\bm{y}|_{2}\leq\epsilon, z_i=y_i, i=1,...,d_1\\
        0,\,\, \text{otherwise},
    \end{cases}
    \label{f_def}
\end{equation}
where $|\bm{z}-\bm{y}|_2\coloneqq\big(\sum_{i=d_1+1}^{d}(z_i-y_i)^2\big)^{\frac{1}{2}}$, $0<\epsilon<<1$ is a small positive number to be determined, and $\phi_{\epsilon}\in C^{\infty}(\mathbb{R}^{d_2}), d_2\coloneqq d-d_1$ is a smooth function with support in $B_0(\epsilon)$ satisfying: 
\begin{equation}    
\int_{\mathbb{R}^{d_2}}\phi_{\epsilon}(\tilde{\bm{x}})\text{d}\tilde{\bm{x}}=1.
\end{equation}
In Eq.~\eqref{f_def}, $\bm{z}\in\mathbb{R}^d$ and $\bm{y}\coloneqq(y_1,...,y_{d_1}, y_{d_1+1},...,y_{d})$ such that $y_i\in\mathbb{N}, i=d_1+1,...,d$. Because $\phi_{\epsilon}$ is a smooth function with compact support, from the last condition in Assumption~\ref{assumptions_w2}, $f_{\epsilon, \bm{x}}(\bm{z})$ is uniformly continuous for $\bm{z}\in\mathbb{R}^d$ for all $\bm{x}\in D$. Furthermore, from the fifth and sixth conditions in Assumption~\ref{assumptions_w2}, it is easy to verify that for every $\bm{x}$, $f_{\epsilon, \bm{x}}$ is a smooth function with compact support in $\mathbb{R}^{d}$ satisfying:
\begin{equation}
         |f_{\epsilon, \bm{x}}|_{\text{mix}_1}\coloneqq \sum_{|\bm{n}|_0\leq d}\|\partial_{\bm{n}}^{|\bm{n}|_0}f_{\bm{x}}\|_{L^2}<\infty,\,\,\,|\sqrt{f_{\epsilon, \bm{x}}}|_{\text{mix}_1}<\infty
         \label{mix1}
     \end{equation}
     where $|\bm{n}|_0$ is the number of nonzero components in $\bm{n}$, $\bm{n}=(n_1,...,n_j)$ satisfying $1\leq n_1<...<n_j\leq d$, and  $\partial_{\bm{n}}f_{\bm{x}}\coloneqq\partial_{y_{n_1}}...\partial_{y_{n_j}}f$. Furthermore, we can verify that 
     \begin{equation}
         |f_{\epsilon, \bm{x}}y_i^2|_{\text{mix}_1}<\infty,\,\,\, |f_{\epsilon, \bm{x}}y_i^2y_j^2|_{\text{mix}_1}<\infty, \,\,\,i, j=1,...,d.
         \label{mix2}
     \end{equation}

For any coupling measure of $(\bm{y}, \tilde{\bm{y}})$ denoted by $\pi_{\bm{x}, \tilde{\bm{x}}}(\cdot, \cdot)$ whose marginal distributions coincide with $f_{\bm{x}}$ and $f_{\tilde{\bm{x}}}$, we can define a new coupling measure:
\begin{equation}
    \pi_{\epsilon, \bm{x}, \tilde{\bm{x}}}(\bm{z}, \tilde{\bm{z}})\coloneqq \begin{cases}
        \pi_{\bm{x}, \tilde{\bm{x}}}(\bm{y}, \tilde{\bm{y}})\delta\big((\bm{z}_2-\bm{y}_2)- (\tilde{\bm{z}}_2-\tilde{\bm{y}}_2)\big)\phi_{\epsilon}(\bm{y}_2-\bm{z}_2),\\\quad\quad \text{if}\,\,\,|\bm{z}-\bm{y}|_{2}<\epsilon, z_i=y_i, \tilde{z}_i=\tilde{y}_i, i=1,...,d_1,\\
        0,\,\, \text{otherwise},
    \end{cases}
\end{equation}
where $\delta(\cdot)$ refers to the Dirac delta function. 
It is easy to verify that the marginal probability distributions of $\pi_{\epsilon, \bm{x}, \tilde{\bm{x}}}(\bm{z}, \tilde{\bm{z}})$ coincide with $f_{\epsilon, \bm{x}}$ and $f_{\epsilon, \tilde{\bm{x}}}$, respectively. Furthermore, we have:
\begin{equation}
\begin{aligned}
        \hat{W}_2(f_{\epsilon, \bm{x}}, f_{\epsilon, \tilde{\bm{x}}})&\leq \inf_{\pi_{\bm{x},\tilde{\bm{x}}}}\,\E_{(\bm{z}, \tilde{\bm{z}})\sim\pi_{\epsilon, \bm{x}, \tilde{\bm{x}}}}\bigg[\sum_{i=1}^d(y_i-\tilde{y}_i)^2 + \sum_{i=d_1+1}^{d}\hat{\delta}_{z_i-\tilde{z}_i, 0}\bigg]^{\frac{1}{2}}\\
        &\quad= \inf_{\pi_{\bm{x},\tilde{\bm{x}}}}\,\E_{(\bm{z}, \tilde{\bm{z}})\sim\pi_{\epsilon, \bm{x}, \tilde{\bm{x}}}}\bigg[\sum_{i=1}^d(y_i-\tilde{y}_i)^2 + \sum_{i=d_1+1}^{d}\hat{\delta}_{y_i-\tilde{y}_i, 0}\bigg]^{\frac{1}{2}}\\
        &= \inf_{\pi_{\bm{x},\tilde{\bm{x}}}}\,\E_{(\bm{y}, \tilde{\bm{y}})\sim\pi_{\bm{x}, \tilde{\bm{x}}}}\big[\|\bm{y}-\tilde{\bm{y}}\|^2\big]^{\frac{1}{2}}= \hat{W}_2(f_{\bm{x}}, f_{\tilde{\bm{x}}}),
\end{aligned}
\end{equation}
where $\hat{\delta}_{z_i-\tilde{z}_i, 0}$ is defined in Eq.~\eqref{delta_function}.
Therefore, $f_{\epsilon, \bm{x}}$ defined in Eq.~\eqref{f_def} also satisfies the Lipschitz condition Eq.~\eqref{l_condition} in Assumption~\ref{assumptions_w2}. From Eq.~\eqref{upper_lower_bounds2}, we also have:
\begin{equation}
    {W}_2(f_{\epsilon, \bm{x}}, f_{\epsilon, \tilde{\bm{x}}})\leq\frac{L}{k}\|\bm{x}-\tilde{\bm{x}}\|_2,
    \label{l_epsilon}
\end{equation}
where $L$ is the Lipschitz constant in Eq.~\eqref{l_condition}.

Combining Eqs.~\eqref{mix1}, \eqref{mix2}, the uniform continuity of $f_{\epsilon, \bm{x}}(\bm{z})$, and the Lipschitz condition Eq.~\eqref{l_epsilon}, the assumptions in the universal approximation ability theorem of SNNs in \cite[Appendix H]{xia2025new} hold. Therefore, for any $\epsilon_0>0$, from \cite[Appendix H]{xia2025new}, there exists an SNN such that:
\begin{equation}
    \int_D W_2^2(f_{\epsilon, \bm{x}}, \hat{f}_{ \bm{x}})\nu(\d\bm{x})<\epsilon_0.
    \label{app_result}
\end{equation}
In Eq.~\eqref{app_result}, $\hat{f}_{\bm{x}}$ refers to the probability measure of the output of the SNN when the input is $\bm{x}$.
(note: \cite[Appendix H]{xia2025new} also imposes some technical regularity conditions on the bounded set $D$ for $\bm{x}$ in Eq.~\eqref{random_field_model}. For simplicity, we assume those conditions hold here.)

 Consider the following coupling measure of $(\bm{y},\tilde{\bm{y}})$: 
\begin{equation}
    \pi_{\epsilon, \bm{x}}(\bm{y},\tilde{\bm{y}})\coloneqq\delta(\bm{y}_1- \tilde{\bm{y}}_1) f_{\epsilon, \bm{x}}(\tilde{\bm{y}}) \mathbf{1}_{|\tilde{\bm{y}}-\bm{y}|_2\leq \epsilon},
\end{equation}
where $\delta(\cdot)$ is the Dirac delta function, $\bm{y}_1$ and $\tilde{\bm{y}}_1$ refers to the first $d_1$ components of $\bm{y}$ and $\tilde{\bm{y}}$, and $\mathbf{1}$ is the indicator function. We can verify that the marginal distributions of $\pi_{\epsilon,\bm{x}}$ coincide with $f_{\bm{x}}$ and $f_{\epsilon, \bm{x}}$. Furthermore, we have:
\begin{equation}
     W_2(f_{\epsilon, \tilde{\bm{x}}}, f_{\tilde{\bm{x}}})\leq \E_{(\bm{y}, \tilde{\bm{y}})\sim\pi_{\epsilon, \bm{x}}}\big[\|\bm{y}-\tilde{\bm{y}}\|^2\big]^{\frac{1}{2}}=\E_{(\bm{y}, \tilde{\bm{y}})\sim\pi_{\epsilon, \bm{x}}}\big[|\bm{y}-\tilde{\bm{y}}|_2^2\big]^{\frac{1}{2}}\leq  \epsilon.
     \label{epsilon_bound}
\end{equation}

Therefore, using Eqs.~\eqref{app_result} and \eqref{epsilon_bound}, we conclude:
\begin{equation}
    \int_D W_2^2(f_{ \bm{x}}, \hat{f}_{ \bm{x}})\nu(\d\bm{x}) \leq 2\int_D W_2^2(f_{\epsilon, \bm{x}}, \hat{f}_{ \bm{x}})\nu(\d\bm{x}) + 2\int_D W_2^2(f_{\epsilon, \tilde{\bm{x}}}, f_{\bm{x}})\nu(\d\bm{x})\leq 2(\epsilon^2+\epsilon_0).
\end{equation}

Applying Eq.~\eqref{upper_lower_bounds}, we conclude that:
\begin{equation}
    \int_D \hat{W}_2^2(f_{\bm{x}}, \hat{f}_{ \bm{x}})\d\bm{x}<2K^2(\epsilon^2+\epsilon_0),
\end{equation}
which proves Theorem~\ref{theorem1} since $\epsilon$ and $\epsilon_0$ can be arbitrarily small.

\section{Proof to Theorem~\ref{theorem_2}}
\label{appendixB}
Here, we provide proof of Theorem~\ref{theorem_2}. First, we have:
    \begin{equation}
    \begin{aligned}
            \E\Big[\big|\hat{W}_2^2(\bm{y}_{\bm{x}}, \hat{\bm{y}}_{\bm{x}}) - \hat{W}_{2, \delta}^{2, \text{e}}(\bm{y}_{\bm{x}}, \hat{\bm{y}}_{\bm{x}})\big|\Big] &\leq \E\Big[\big|\hat{W}_2^2(\bm{y}_{\bm{x}}, \hat{\bm{y}}_{\bm{x}}) - \hat{W}_{2}^{2, \text{e}}(\bm{y}_{\bm{x}}, \hat{\bm{y}}_{\bm{x}})\big|\Big] \\
            &\hspace{1cm}+ \E\Big[\big|\hat{W}_2^{2, \text{e}}(\bm{y}_{\bm{x}}, \hat{\bm{y}}_{\bm{x}}) - \hat{W}_{2, \delta}^{2, \text{e}}(\bm{y}_{\bm{x}}, \hat{\bm{y}}_{\bm{x}})\big|\Big],
            \end{aligned}
            \label{estimate_term}
    \end{equation}
    where 
\begin{equation}
    \hat{W}_{2}^{2, \text{e}}(\bm{y}_{\bm{x}}, \hat{\bm{y}}_{\bm{x}})\coloneqq \int_{D}\hat{W}_2^2\big(f_{\bm{x}}, \hat{f}_{\bm{x}}\big)\nu^{\text{e}}(\text{d}\bm{x}),
\end{equation}
and $\nu(\d\bm{x}), \nu^{\text{e}}(\d\bm{x})$ are the probability measure and empirical probability measure of $\bm{x}$, respectively.

For the first term in Eq.~\eqref{estimate_term}, the following inequality holds:
\begin{equation}
\begin{aligned}
&\E\Big[\big|\int_{D}\hat{W}_2^2\big(f_{\bm{x}}, \hat{f}_{\bm{x}}\big)\nu^{\text{e}}(\text{d}\bm{x}) - \int_{D}\hat{W}_2^2\big(f_{\bm{x}}, \hat{f}_{\bm{x}}\big)\nu(\text{d}\bm{x})\big|\Big]\\
    &\hspace{1cm}\leq \E\Big[\big(\int_{D}\hat{W}_2^2\big(f_{\bm{x}}, \hat{f}_{\bm{x}}\big)\nu^{\text{e}}(\text{d}\bm{x}) - \int_{D}\hat{W}_2^2\big(f_{\bm{x}},\hat{f}_{\bm{x}}\big)\nu(\text{d}\bm{x})\big)^2\Big]^{\frac{1}{2}}\\
    &\hspace{2cm}\leq \frac{1}{\sqrt{N}}\E\bigg[\Big(\hat{W}_2^2\big(f_{\bm{x}}, \hat{f}_{\bm{x}}\big) - \E[\hat{W}_2^2\big(f_{\bm{x}}, \hat{f}_{\bm{x}}\big)]\Big)^2\bigg]^{\frac{1}{2}}
    \leq \frac{4M}{\sqrt{N}}.
    \label{intermediate1}
\end{aligned}
\end{equation}
The last inequality holds because for any $\bm{x}\in D$, using the assumption Eq.~\eqref{upperboundy}, we have
\begin{equation}
   0\leq \hat{W}_2^2\big(f_{\bm{x}}, \hat{f}_{\bm{x}}\big) \leq 2 \Big(\E\big[\|\bm{y}_{\bm{x}}\|^2\big] + \E\big[\|\hat{\bm{y}}_{\bm{x}}\|^2\big]\Big)=4M.
\end{equation}

Next, we estimate the second term in Eq.~\eqref{estimate_term}:
\begin{equation}
    \E\bigg[\Big|\int_{D}\hat{W}_2^2\big(f_{\bm{x}}, \hat{f}_{\bm{x}}\big)\nu^{\text{e}}(\text{d}\bm{x}) - \int_{D}\hat{W}_2^2\big(f_{\bm{x}, \delta}^{\text{e}}, \hat{f}_{\bm{x}, \delta}^{\text{e}}\big)\nu^{\text{e}}(\text{d}\bm{x})\Big|\bigg].
\end{equation}
We denote $f_{\bm{x}, \delta}$ and $f_{\bm{x}, \delta}^{\text{e}}$ to be the conditional probability measure and the empirical conditional measure of $\bm{y}_{\tilde{\bm{x}}}$ conditioned on $\|\tilde{\bm{x}}-\bm{x}\|_2\leq \delta$. Similarly, we denote $\hat{f}_{\bm{x}, \delta}$ and $\hat{f}_{\bm{x}, \delta}^{\text{e}}$ to be the conditional distribution and the empirical conditional distribution of $\hat{\bm{y}}_{\tilde{\bm{x}}}$
 conditioned on $\|\tilde{\bm{x}}-\bm{x}\|_2\leq \delta$, respectively.

For any $\bm{x}\in D$, we have
\begin{equation}
\begin{aligned}
&\hspace{-1cm}\Big|\hat{W}_2^2\big(f_{\bm{x}}, \hat{f}_{\bm{x}}\big) - \hat{W}_2^2\big(f_{\bm{x}, \delta}^{\text{e}}, \hat{f}_{\bm{x}, \delta}^{\text{e}}\big)\Big|\leq  \bigg(\hat{W}_2\big(f_{\bm{x}}, \hat{f}_{\bm{x}}\big) + \hat{W}_2\big(f_{\bm{x}, \delta}^{\text{e}}, \hat{f}_{\bm{x}, \delta}^{\text{e}}\big)\bigg) \\&
\hspace{5cm}\cdot|\hat{W}_2\big(f_{\bm{x}}, \hat{f}_{\bm{x}}\big) - \hat{W}_2\big(f_{\bm{x}, \delta}^{\text{e}}, \hat{f}_{\bm{x}, \delta}^{\text{e}}\big)|
\\&\hspace{4.2cm}\leq4\sqrt{M}\Big|\hat{W}_2\big(f_{\bm{x}}, \hat{f}_{\bm{x}}\big) - \hat{W}_2\big(f_{\bm{x}, \delta}^{\text{e}}, \hat{f}_{\bm{x}, \delta}^{\text{e}}\big)\Big|.
\end{aligned}
    \label{break_ineq}
\end{equation}

Using the triangle inequality of the Wasserstein distance in \cite[Proposition 2.1]{clement2008elementary}, for any $\bm{x}$, we have
\begin{equation}
    \begin{aligned}
        &\hspace{0cm}\big|\hat{W}_2(f_{\bm{x}}, \hat{f}_{\bm{x}}) - \hat{W}_2(f_{\bm{x}, \delta}^{\text{e}}, \hat{f}_{\bm{x},  \delta}^{\text{e}}) \big|\leq \big|\hat{W}_2(\hat{f}_{\bm{x}}, f_{\bm{x}}) - \hat{W}_2(\hat{f}_{\bm{x}}, f_{\bm{x}, \delta})\big| + \big|\hat{W}_2(\hat{f}_{\bm{x}}, f_{\bm{x}, \delta}) - \hat{W}_2(f_{\bm{x}, \delta}, \hat{f}_{\bm{x}, \delta})\big| \\
    &\hspace{4cm}+ \big|\hat{W}_2(f_{\bm{x}, \delta}, \hat{f}_{\bm{x}, \delta}) - \hat{W}_2(\hat{f}_{\bm{x}, \delta}, f_{\bm{x}, \delta}^{\text{e}})\big|  + \big|\hat{W}_2(\hat{f}_{\bm{x}, \delta}, f_{\bm{x}, \delta}^{\text{e}}) - \hat{W}_2(f_{\bm{x}, \delta}^{\text{e}}, \hat{f}^{\text{e}}_{\bm{x}, \delta})\big|\\
    &\hspace{4cm}\leq \hat{W}_2(f_{\bm{x}, \delta}, {f}_{\bm{x}}) +  \hat{W}_2(\hat{f}_{\bm{x}, \delta}, \hat{f}_{\bm{x}}) + \hat{W}_2(f_{\bm{x},\delta}^{\text{e}}, f_{\bm{x}, \delta}) + \hat{W}_2(\hat{f}^{\text{e}}_{\bm{x}, \delta}, \hat{f}_{\bm{x}, \delta})\\
    &\hspace{3.5cm}\leq \hat{W}_2(f_{\bm{x}, \delta}, {f}_{\bm{x}}) +  \hat{W}_2(\hat{f}_{\bm{x}, \delta}, \hat{f}_{\bm{x}}) + KW_2(f_{\bm{x},\delta}^{\text{e}}, f_{\bm{x}, \delta}) + KW_2(\hat{f}^{\text{e}}_{\bm{x}, \delta}, \hat{f}_{\bm{x}, \delta})
    \label{misalign}
    \end{aligned}
\end{equation}

For any $\epsilon_2>0$, using the Lipschitz condition Eq.~\eqref{l_condition} in Assumption~\ref{assumptions_w2}, there exists a coupling measure denoted by $\pi_{\bm{x}, \tilde{\bm{x}}, \epsilon_2}$ whose marginal distributions coincide with $f_{\bm{x}}$ and $f_{\tilde{\bm{x}}}$ satisfying:
\begin{equation}
    \E_{(\bm{y}, \tilde{\bm{y}})\sim \pi_{\bm{x}, \tilde{\bm{x}}, \epsilon_2}}\big[\|\bm{y}-\tilde{\bm{y}}\|^2\big]\leq \hat{W}_2^2(f_{\bm{x}}, f_{\tilde{\bm{x}}})+\epsilon_2\leq L^2\|\bm{x}-\tilde{\bm{x}}\|_2^2+\epsilon_2.
\end{equation}
 Consider a special coupling measure of $(\bm{y}, \tilde{\bm{y}})$ defined as:
\begin{equation}
    \pi_{\bm{x}, \delta, \epsilon_2}(\bm{y},\tilde{\bm{y}})\coloneqq \int_{B(\bm{x},\delta)}\pi_{\bm{x}, \tilde{\bm{x}}, \epsilon_2}(\bm{y}, \tilde{\bm{y}})\frac{\nu(\d\tilde{\bm{x}})}{\nu(B(\bm{x},\delta))},
\end{equation}
where $B(\bm{x},\delta)$ is the ball $\{\tilde{x}\in D:\|\tilde{\bm{x}}-\bm{x}\|_2\leq\delta\}$.
We can verify that the marginal distributions of $\pi_{\bm{x}, \delta, \epsilon_2}(\bm{y},\tilde{\bm{y}})$ coincide with $f_{\bm{x}}$ and $f_{\bm{x}, \delta}=\frac{1}{B(\bm{x},\delta)}\int_{B(\bm{x},\delta)}f_{\tilde{\bm{x}}}\nu(\d\tilde{\bm{x}})$. Furthermore, we have:
\begin{equation}
\begin{aligned}
        \hat{W}_2^2(f_{\bm{x}, \delta}, {f}_{\bm{x}})&\leq \E_{(\bm{y}, \tilde{\bm{y}})\sim \pi_{\bm{x}, \delta, \epsilon_2}}\big[\|\bm{y}-\tilde{\bm{y}}\|^2\big]\leq \frac{1}{B(\bm{x},\delta)}\int_{B(\bm{x},\delta)}\E_{(\bm{y}, \tilde{\bm{y}})\sim\pi_{\bm{x}, \tilde{\bm{x}}, \epsilon_2}}\big[\|\bm{y}-\tilde{\bm{y}}\|^2\big]\nu(\d\tilde{\bm{x}})\\
        &\quad\leq L^2\delta^2 + \epsilon_2
\end{aligned}
\end{equation}


For the third and fourth terms of the last inequality of Eq.~\eqref{misalign}, from Theorem 1 in \cite{fournier2015rate}, there exists a constant $C$ such that:
\begin{equation}
    \E\Big[W_2(f_{\bm{x},\delta}^{\text{e}}, f_{\bm{x},\delta})\Big]\leq \E\Big[W_2^2(f_{\bm{x},\delta}^{\text{e}}, f_{\bm{x},\delta})\Big]^{\frac{1}{2}}\leq C\E\Big[\|\bm{y}_{\bm{x}}\|_6^6\Big]^{\frac{1}{6}}h(N(x, \delta),{d})\leq C\sqrt{M}h\big(N(\bm{x}, \delta),{d}\big)
    \label{sample_bound}
\end{equation}
and 
\begin{equation}
\E\Big[W_2(\hat{f}_{\bm{x},\delta}^{\text{e}}, \hat{f}_{\bm{x},\delta})\Big]\leq \E\Big[W_2^2(\hat{f}_{\bm{x},\delta}^{\text{e}}, \hat{f}_{\bm{x},\delta})\Big]^{\frac{1}{2}}\leq C\E\Big[\|\hat{\bm{y}}_{\bm{x}}\|_6^6\Big]^{\frac{1}{6}}h(N(\bm{x}, \delta), {d})\leq C\sqrt{M}h\big(N(\bm{x}, \delta),{d}\big),
    \label{sample_bound2}
\end{equation}
respectively. In Eq.~\eqref{sample_bound2}, $\|\cdot\|_6$ is the $\ell^6$ norm of a vector in $\mathbb{R}^d$, and we have $\|\bm{y}\|_6\leq\|\bm{y}\|_2$. In Eq.~\eqref{sample_bound2}, the function $h$ is defined as:
\begin{equation}
h(N, {d})=\left\{
\begin{aligned}
&N^{-\frac{1}{4}}\log(1+N)^{\frac{1}{2}}, d\leq4,\\
&N^{-\frac{1}{{d}}}, {d}> 4.
\end{aligned}
\right.
\end{equation}
Therefore, we conclude that:
\begin{equation}
    \E\Big[\big|\hat{W}_{2}^{2, \text{e}}(\bm{y}_{\bm{x}}, \hat{\bm{y}}_{\bm{x}}) - \hat{W}_{2, \delta}^{2, \text{e}}(\bm{y}_{\bm{x}}, \hat{\bm{y}}_{\bm{x}})\big|\Big]\leq 8KCM\E\big[h(N(\bm{x}, \delta), {d})\big] + 8\sqrt{M}\sqrt{L^2\delta^2+\epsilon_2}.
    \label{intermediate2}
\end{equation}
Combining the two inequalities Eqs.~\eqref{intermediate1} and \eqref{intermediate2}, the inequality~\eqref{theorem2_result} holds because $\epsilon_2$ can be chosen to be arbitrarily small, completing the proof of Theorem~\ref{theorem_2}. 

\section{Default training settings and hyperparameters}
\label{appendix_training}
We list the hyperparameters and settings for training the SNN model in Fig.~\ref{fig:snn} of each example in Table~\ref{tab:setting}.
\begin{table}[h!]
\tiny
\centering
\caption{Training settings and hyperparameters for each example.} 
\begin{tabular}{lllll}
\toprule {Hyperparameters} & Example \ref{example1} & Example \ref{example2} &
Example \ref{example3} &
Example \ref{example4} \\
\midrule
Gradient descent method & Adam & Adam & Adam & Adam \\
Learning rate & 0.001 & 0.01 & 0.005 & 0.01\\
Weight decay & 0.01 & $10^{-4}$ & $10^{-4}$ & 0\\
\# of epochs $\text{epoch}_{\max}$& 3000 & 2000 & 1000 & 1000\\
$\lambda$ in Eq.~\eqref{w2_distance} &$\backslash$ &$\backslash$ & $\text{Var}[y_1]$& $\frac{1}{T+1}\sum_{j=0}^T\sum_{i=1}^2\big(\text{Var}[m_i(t_j)] $\\
& & & & $+ \text{Var}[p_i(t_j)] + \text{Var}[d_i(t_j)]\big)$ \\
\# of training samples $N$ & 1000 & {4000} & 3341 & 300 \\
Hidden layers & 5 & 5 & 5 & 5\\
\# of epochs to update the minibatch $\text{epoch}_{\text{update}}$ & 50 & 50 &50 & $\backslash$\\
\# of data points in a minibatch $n$ & 100 & 1000 & 300 & 300\\
neighborhood size $\delta$ &0.025 & $0.5\sqrt{8}$& $0.3\sqrt{7}$ &0.02 \\
Activation function & {GELU} & {GELU} & {GELU} & GELU\\
Equipped with the ResNet technique? & Yes & Yes & Yes & Yes\\
Neurons in each layer  & 32 & 32 & 32 & 16 \\
Initialization for biases &  $\mathcal{N}(0, 0.05^2)$ &  $\mathcal{N}(0, 0.05^2)$
&  $\mathcal{N}(0, 0.05^2)$ &  $\mathcal{N}(0, 0.05^2)$ \\ 
Initialization for the means of weights &  $\mathcal{N}(0, 0.05^2)$ & $\mathcal{N}(0, 0.05^2)$ & $\mathcal{N}(0, 0.05^2)$ &  $\mathcal{N}(0, 0.05^2)$\\
Initialization for the variances of weights &  $\mathcal{N}(0, 0.05^2)$ & $\mathcal{N}(0, 0.05^2)$ & $\mathcal{N}(0, 0.05^2)$ &  $\mathcal{N}(0, 0.05^2)$
  \\
\bottomrule
\end{tabular}
\label{tab:setting}
\end{table}

\section{Sensitivity tests of Example~\ref{example1}}
\label{example1_appendix}
\begin{figure}
    \centering
\includegraphics[width=0.95\linewidth]{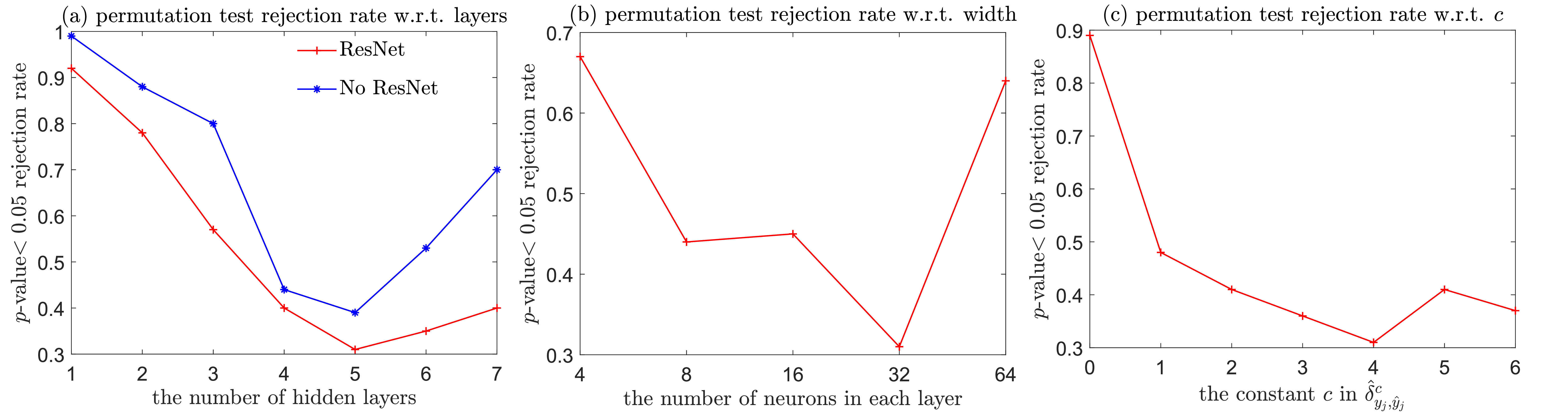}
    \caption{(a) the $p$-value test rejection rate w.r.t. the number of hidden layers in the SNN. The forward propagation mode is either the Normal mode or the ResNet model, as described in Fig.~\ref{fig:snn}. All hidden layers have 32 neurons. (b) the $p$-value test rejection rate w.r.t. the number of neurons in each hidden layer in SNNs. All SNNs have 5 hidden layers and adopt the ResNet technique for forward propagation. (c) the $p$-value test rejection rate w.r.t. $c$ in Eq.~\eqref{delta_function_revised}.}
    \label{fig:example1_appendix}
\end{figure}

In this section, we carry out additional sensitivity tests for Example~\ref{example1}. First, we investigate how the architecture of the SNN, \textit{i.e.} the number of neurons in each layer, the number of hidden layers in the SNN model (Fig.~\ref{fig:snn}), as well as whether adopting the ResNet technique \cite{he2016deep} for forward propagation would affect the accuracy of the reconstructed random field model in Example~\ref{example1}. We set $\sigma=0.4$ in Eq.~\eqref{example1_model} and all training settings and hyperparameters are the same as Example~\ref{example1}, which is shown in Table~\ref{tab:setting}.
We use the $p$-value test rejection rate on the same testing set as used in Example~\ref{example1} to evaluate how SNNs with different structures can reconstruct the random field model Eq.~\eqref{example1_model}. The results are shown in Fig.~\ref{fig:example1_appendix} (a)(b) and Table~\ref{tab:nn_structure}.

\begin{table}[h!]
\scriptsize
\centering
  \caption{The $p$-value test rejection rate of the reconstructed random field model for Example~\ref{example1}. The ResNet technique is used for forward propagation.}
\begin{tabular}{ccccc}
\toprule  width & \# of layers & activation function & rejection rate   \\ 
\midrule 
  32 & 5 & GELU &  0.31
 \\ 
    32 & 5 & ReLU &   0.83
 \\ 
     32 & 5 & ELU ($\alpha=1$) &   0.52
 \\ 
     32 & 5 & Leaky ReLU (0.01) & 0.49
 \\ 
\bottomrule
\end{tabular}
\label{tab:nn_structure}
\end{table}

From Fig.~\ref{fig:example1_appendix} (a)(b), SNNs with a too small number of hidden layers or too few neurons in each layer are incapable of accurately reconstructing the model Eq.~\eqref{example1_model}. On the other hand, SNNs with more than 5 hidden layers or more than 32 neurons in each layer yield worse performance compared to the SNN with 5 hidden layers and 32 neurons in each layer. This indicates that the training of a deeper or wider SNN could be more complicated and requires more tuning of hyperparameters.
Additionally, as shown in Fig.~\ref{fig:example1_appendix} (a),
the ResNet technique can improve SNNs' capability for approximating the random field model Eq.~\eqref{example1_model}. Finally, among all activation functions, we find that using the GELU activation function gives the most accurate reconstruction of Eq.~\eqref{example1_model} 
(shown in Table~\ref{tab:nn_structure}).

Next, we replace the constant 4 in Eq.~\eqref{delta_function} with other constants, \textit{i.e.}, replacing $\hat{\delta}_{y_j, \hat{y}_j}$ in Eq.~\eqref{w2_distance} with:
\begin{equation}
\begin{aligned}
        \hat{\delta}_{y_j, \hat{y}_j}^{c}=\begin{cases}
        c(y_j-\hat{y}_j)^2, |y_j- \hat{y}_j|\leq \frac{1}{2},\\
        1,\,\, |y_j- \hat{y}_j|>\frac{1}{2}.
    \end{cases}
\end{aligned}
\label{delta_function_revised}
\end{equation}
We reconstruct the model Eq.~\eqref{example1_model} by varying $c$ in Eq.~\eqref{delta_function_revised} and record the $p$-value test rejection rate on the same testing set as used in Example~\ref{example1}. Setting $c$ in Eq.~\eqref{delta_function_revised} to be too small or too large will both result in less accurate reconstruction of Eq.~\eqref{example1_model}, and $c=4$ seems to be the most appropriate choice for an accurate reconstruction of Eq.~\eqref{example1_model} (shown in Fig.~\ref{fig:example1_appendix} (c)).

\bibliographystyle{siam}
\bibliography{ref.bib}
\end{document}